\documentclass[11pt]{article}

\usepackage[preprint]{acl}

\usepackage{times}
\usepackage{latexsym}

\usepackage[T1]{fontenc}
\usepackage[utf8]{inputenc}

\usepackage{microtype}

\usepackage{inconsolata}

\usepackage{graphicx}

\usepackage{hyperref}
\usepackage{url}
\usepackage{subfig}
\usepackage{xspace}
\usepackage[table]{xcolor}
\usepackage{booktabs}
\usepackage{enumitem}
\usepackage{multirow}
\usepackage{amsfonts}
\usepackage{amsmath} 
\usepackage{makecell}
\usepackage{wrapfig}
\usepackage{longtable}
\usepackage{subscript}
\usepackage{array}

\usepackage{xcolor}    
\usepackage{tcolorbox}

\title{ZARA: Training-Free Motion Time-Series Reasoning via Evidence-Grounded LLM Agents}

\author{
Zechen Li$^{1}$ \quad Baiyu Chen$^{1}$ \quad Hao Xue$^{1,2,3}$ \quad Flora D. Salim$^{1}$ \vspace{1mm}\\
$^{1}$University of New South Wales, Sydney \\
$^{2}$Hong Kong University of Science and Technology (Guangzhou) \\
$^{3}$Hong Kong University of Science and Technology \vspace{1mm} \\
\texttt{\{zechen.li, breeze.chen, flora.salim\}@unsw.edu.au \quad haoxue@hkust-gz.edu.cn}
}

\begin{document}
\maketitle
\begin{abstract}
Motion sensor time-series are central to Human Activity Recognition (HAR), yet conventional approaches are constrained to fixed activity sets and typically require costly parameter retraining to adapt to new behaviors. While Large Language Models (LLMs) offer promising open-set reasoning capabilities, applying them directly to numerical time-series often leads to hallucinations and weak grounding. To address this challenge, we propose \textbf{ZARA} (\underline{Z}ero-training \underline{A}ctivity \underline{R}easoning \underline{A}gents), a knowledge- and retrieval-augmented agentic framework for motion time-series reasoning in a training-free inference setting. Rather than relying on black-box projections, ZARA distills reference data into a statistically grounded textual knowledge base that transforms implicit signal patterns into verifiable natural-language priors. Guided by retrieved evidence, ZARA iteratively selects discriminative cues and performs grounded reasoning over candidate activities. Extensive experiments on eight benchmarks show that ZARA generalizes robustly to unseen subjects and across datasets, demonstrating strong transferability across heterogeneous sensor domains. These results mark a step toward trustworthy, plug-and-play motion understanding beyond dataset-specific artifacts. Our code is available at \url{https://github.com/zechenli03/ZARA}.
\end{abstract}

%%%%%%%%% BODY TEXT
\section{Introduction}
\label{introduction}

\begin{figure}[t]
\centering
\includegraphics[width=1\linewidth]{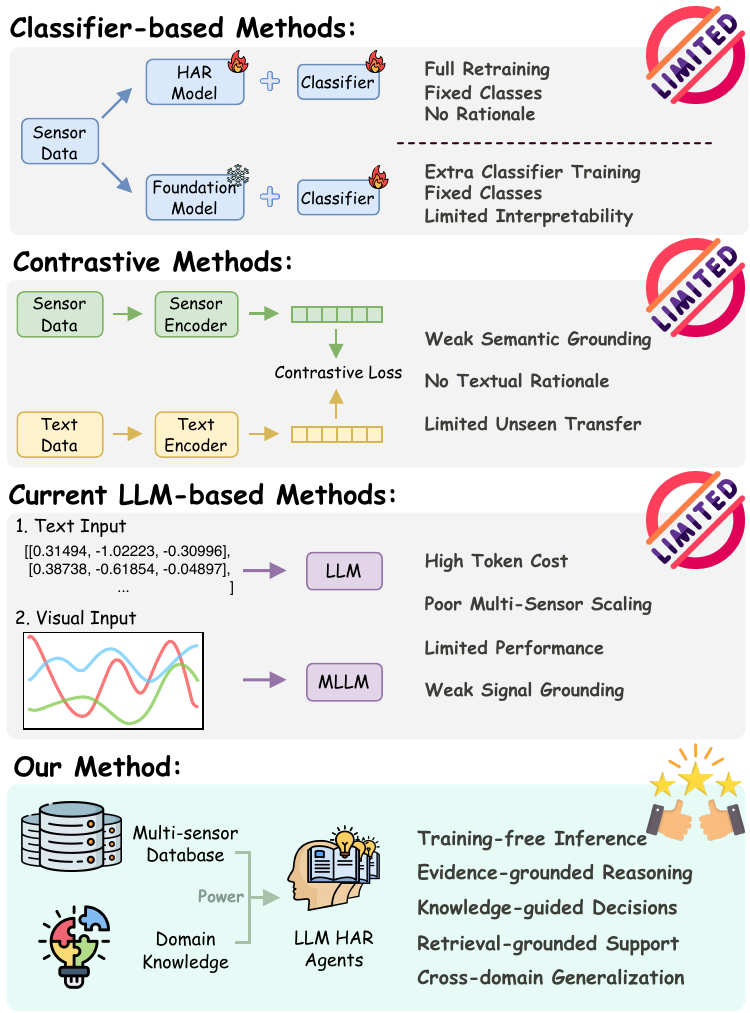} 
\caption{Representative method families for human activity recognition.}
\label{fig:compare}
\vspace{-0.5em}
\end{figure}

Human activity recognition (HAR) from on-body motion sensors underpins applications from digital health to adaptive interfaces. However, the dominant paradigm in HAR remains heavily supervised. Most existing systems typically rely on task-specific deep neural networks~\cite{s16010115, 10.1145/3448083, NIPS2017_3f5ee243} optimized for fixed sensor setups and classes. 

Consequently, existing HAR methods (see Figure~\ref{fig:compare}) face three critical barriers to scalable deployment. 
\textbf{Poor Generalization.} Adapting to new users (cross-subject) or hardware setups (cross-domain) typically necessitates costly model parameter optimization.
\textbf{Limited Training-Free Adaptation.} Time-series (TS) foundation models like Moment~\cite{goswami2024moment} and Mantis~\cite{feofanov2025mantislightweightcalibratedfoundation} offer transferable representations but still require task-specific classification heads. Contrastive models such as UniMTS~\cite{zhang2024unimtsunifiedpretrainingmotion} eliminate the classifier, yet still struggle to distinguish fine-grained activities in parameter-frozen settings due to limited semantic grounding.
\textbf{Lack of Interpretability.} Most approaches yield only categorical predictions without transparent reasoning, limiting trust in safety-critical scenarios.

Meanwhile, LLMs enhanced with retrieval-augmented generation (RAG) have shown strong reasoning capabilities in vision and NLP~\cite{baek-etal-2023-knowledge-augmented, NEURIPS2022_63fef080, 10205444, zhang2025websearchagenticdeep}. However, sensor-based HAR has largely failed to capitalize on this paradigm. Early attempts to apply LLMs to HAR have focused on converting multi-channel signals into token sequences or images. These modality-projection approaches suffer from \textbf{excessive token usage}, \textbf{significant information loss} during discretization, and \textbf{mediocre accuracy}, as LLMs struggle to intuit physical dynamics directly from raw numerical streams.

We argue that the missing link lies in translating implicit signal statistics into explicit, structured language. Just as RAG in NLP relies on a high-quality document corpus, RAG in HAR requires a domain-specific knowledge base that articulates \textit{how} physical movements manifest in sensor data. When equipped with (i) statistically grounded textual priors (e.g., running exhibits higher vertical acceleration variance than walking) and (ii) a retrieval mechanism for relevant signal evidence, LLMs can perform robust reasoning on HAR tasks. This enables evidence-grounded, training-free inference, effectively replacing task-specific classifier training with in-context conditioning on retrieved priors.

Motivated by this insight, we introduce ZARA, a novel agentic framework for HAR in a training-free inference setting via knowledge- and retrieval-augmented reasoning. ZARA bridges the signal-to-language gap via three synergistic components. First, \textbf{Offline Statistical Profiling} automatically distills a general-purpose knowledge base from every activity pair by extracting discriminative feature profiles. This pairwise formulation translates implicit signal characteristics into verifiable linguistic priors, enabling the system to accommodate new activities by simply registering their profiles without parameter updates.
Second, \textbf{Class-Wise Multi-Sensor Retrieval} fetches top-$k$ evidence conditionally per class from a labeled support set to ensure balanced recall across long-tail classes, and then aggregates the heterogeneous sensor-specific rankings using Reciprocal Rank Fusion~\cite{10.1145/1571941.1572114}.
Finally, \textbf{Hierarchical Multi-Agent Reasoning} orchestrates specialized LLM agents to iteratively filter features and prune candidates, progressively narrowing the hypothesis space to produce predictions supported by human-readable explanations.

We benchmark ZARA against 10 established baselines across 8 diverse HAR datasets. Our comprehensive evaluation spans both \textit{Cross-Subject} scenarios (addressing new user adaptation) and \textit{Cross-Dataset} scenarios (testing domain generalization). By fusing structured sensor knowledge with LLM-based reasoning, ZARA offers a plug-and-play alternative to training-intensive pipelines. Concretely, this work contributes:
\begin{itemize}
    \item \textbf{Signal-to-Text Knowledge Grounding.} We propose an automated method to distill motion TS into a pairwise textual knowledge base, enabling LLMs to perform verifiable reasoning in a parameter-frozen setting.
    \item \textbf{Agentic Framework for Interpretable HAR.} ZARA is the first knowledge- and retrieval-driven agentic system for multi-sensor TS classification that also generates concise, evidence-backed rationales, enhancing trust in automated decision-making.
    \item \textbf{Strong Training-Free Generalization.} Extensive experiments confirm ZARA's state-of-the-art performance in parameter-frozen settings, where robust generalization across unseen subjects and heterogeneous domains demonstrates the transferability of the injected motion priors.
\end{itemize}
\section{Related Work}
\label{related_work}
Recent HAR work has shifted from training task-specific networks to using pre-trained foundation models for general time-series (TS) representations. Chronos~\cite{ansari2024chronos} tokenizes TS via scaling and quantization to enable text-style encoder–decoder training; Moment~\cite{goswami2024moment} pre-trains transformers with masked-value prediction; and Mantis~\cite{feofanov2025mantislightweightcalibratedfoundation} uses a contrastively pre-trained Vision Transformer~\cite{dosovitskiy2020image} tailored for time-series classification. Despite strong representations, these backbones still typically require training downstream classifiers for specific tasks.

\begin{figure*}[t]
\centering
\includegraphics[width=1\textwidth]{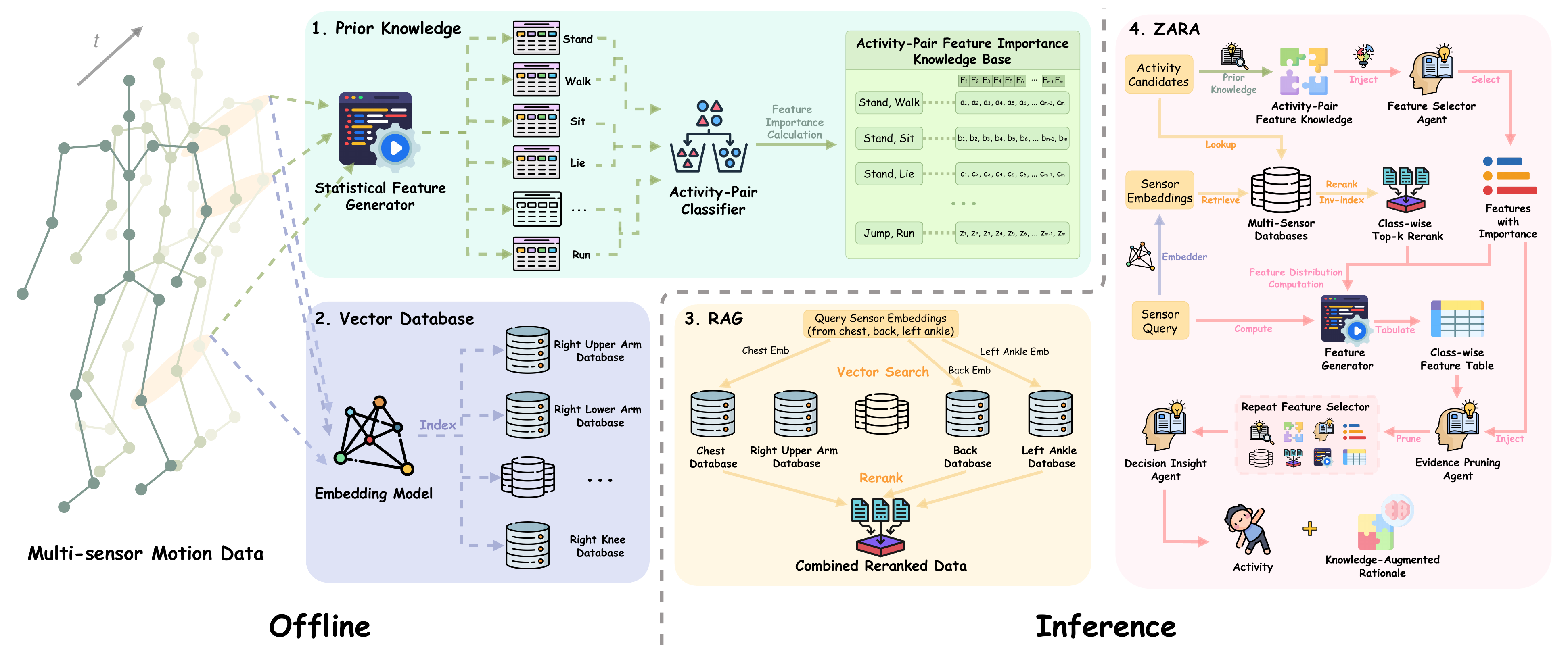} 
\caption{Overall architecture of ZARA, an evidence-grounded agentic framework augmented with knowledge and retrieval for motion time-series reasoning.}
\label{fig:zara}
\vspace{-0.5em}
\end{figure*}

To enable parameter-frozen HAR, prior work explores cross-modal alignment, mapping motion signals into shared embedding spaces with text or images. ImageBind~\cite{girdhar2023imagebind} and IMU2CLIP~\cite{moon-etal-2023-imu2clip} align IMU data with VLM spaces~\cite{pmlr-v139-radford21a}. UniMTS~\cite{zhang2024unimtsunifiedpretrainingmotion} aligns synthetic skeleton motions with text for classifier-free recognition, but often lacks semantic granularity for complex activities. 
COMODO~\cite{chen2025comodo} distills semantics from paired video--IMU data via cross-modal self-supervision, but still relies on task-specific training.

In parallel, recent work has explored LLM agents that leverage long-context memory and external context at test time~\cite{zhang2025personaagentlargelanguagemodel, zhang2026memorycdbenchmarkinglongcontextuser}, while LLM-based reasoning methods apply similar ideas to HAR. HARGPT~\cite{ji2024hargptllmszeroshothuman} and \citet{yoon-etal-2024-eyes} prompt on raw signals via chain-of-thought or visual transformations, incurring high token cost and information loss. SensorLLM~\cite{li-etal-2025-sensorllm} generates human-readable captions but relies on fine-tuning. ZeroHAR~\cite{Chowdhury_Kapila_Panse_Zhang_Teng_Kulkarni_Hong_Gupta_Shang_2025} and SensorLM~\cite{zhang2025sensorlmlearninglanguagewearable} add spatial metadata or hierarchical captions, but still lack verifiable, statistically grounded motion knowledge for robust training-free generalization.
\section{Methodology}
\label{methods}

\begin{figure*}[t]
\centering
\includegraphics[width=1\textwidth]{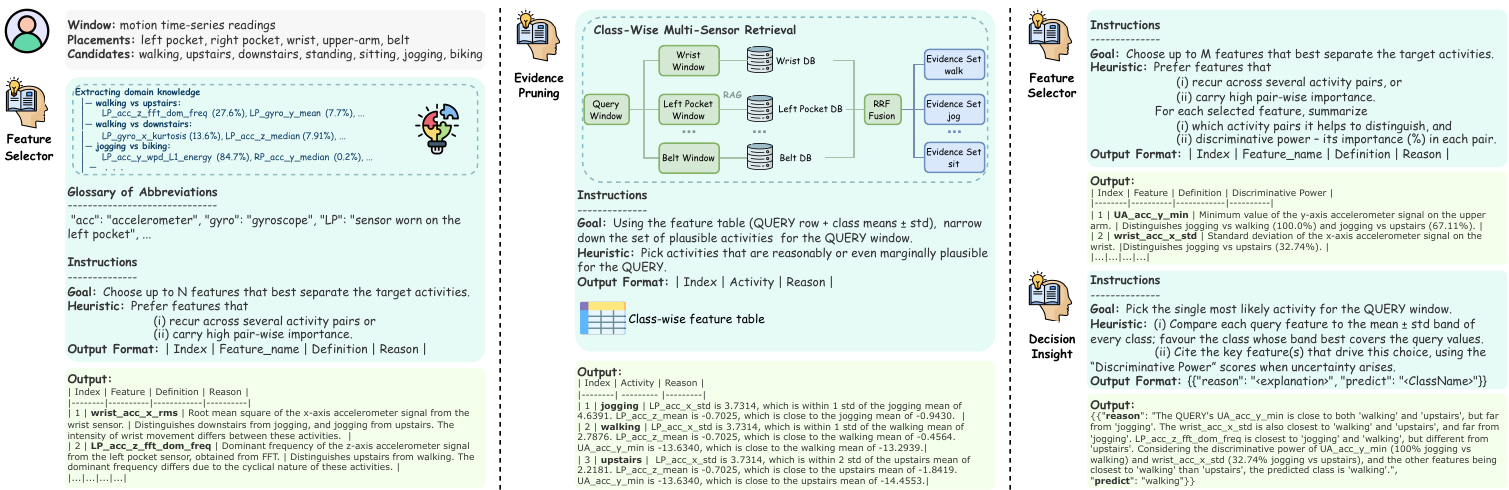} 
\caption{ZARA's multi-agent workflow with placement-specific, class-wise evidence retrieval and rank fusion.}
\label{fig:prompt_flow}
% \vspace{-0.5em}
\end{figure*}

Figure~\ref{fig:zara} illustrates the overall framework of ZARA. We first motivate the decoupling of universal knowledge and local evidence, and then describe the construction of the knowledge base, retrieval backbone, and hierarchical agentic workflow.

\paragraph{Decoupling Knowledge and Evidence.}
Standard RAG systems typically retrieve raw samples directly. However, raw sensor signals lack explicit semantic structure, making it difficult for LLMs to reason about fine-grained physical differences. ZARA addresses this by decoupling information into two sources:
\textit{Universal Knowledge ($\mathcal{K}$):} A static reference registry storing \textit{pairwise feature-importance profiles}. 
Rather than embedding sensor-grounded priors into model weights, it acts as a lookup table identifying which physical properties are most discriminative for separating specific activities (e.g., instructing the agent that "vertical acceleration variance" is the critical metric to distinguish Running from Walking).
\textit{Local Evidence ($\mathcal{D}$):} A vector database of raw-signal embeddings that serves as external memory. This provides \textit{local distributional grounding}, allowing the model to adapt to specific sensor placements or users via in-context retrieval rather than weight adaptation.

\paragraph{Offline Statistical Profiling (Global Priors).}
To equip the LLM with structured, sensor-specific priors, we automatically construct a pairwise Activity Feature Importance Knowledge Base $\mathcal{K}$ through offline statistical analysis. Each wearable unit provides raw sensor channels (typically a 3-axis accelerometer and/or gyroscope). For every labelled window $x_a\!\in\!\mathbb{R}^{T\times C}$ of activity $a$ ($T$ time steps, $C$ channels), we derive a feature pool $\mathcal{F}$ comprising low-cost, human-interpretable statistics (see Appendix~\ref{apd_feats}): \textit{time-domain} measures (mean, variance, RMS, etc.), \textit{frequency-domain} descriptors (spectral entropy, dominant frequency, etc.), and \textit{cross-channel} indicators (correlations, tilt angles). 
For each ordered activity pair $(a_i,a_j)$, we estimate an importance score $s[f,(a_i,a_j)]$ for every $f\!\in\!\mathcal{F}$ using permutation-based feature ranking via AutoGluon~\cite{agtabular}. Cross-validation with fold-weighted averaging yields robust estimates that generalize across folds. All feature–score tuples are stored as $\mathcal{K}[(a_i,a_j)] = [(f_1,s_1),\dots,(f_P,s_P)]$. Because $\mathcal{K}$ is organized pairwise, it translates implicit signal characteristics into verifiable linguistic priors that can be dynamically instantiated for any candidate subset at inference time. Crucially, adding a new activity requires only registering its statistical profile against existing classes, thereby eliminating the need for task-specific retraining or manual rule curation.

\paragraph{Placement-Specific Retrieval (Local Evidence).}
To ensure retrieved evidence aligns with the query’s physical context, we maintain a set of placement-specific vector stores $\{\mathcal{D}^{\text{loc}}\}$, where \textit{loc} denotes the sensor position (e.g., wrist, ankle). Each database acts as a distributional anchor, indexing historical motion windows embedded by a frozen TS foundation encoder $g(\cdot)$ (Mantis~\cite{feofanov2025mantislightweightcalibratedfoundation} by default), stored alongside their statistical features, labels, and sensor metadata. The resulting embedding vectors are L2-normalized and indexed using FAISS IndexFlatIP~\cite{douze2024faiss}. For a query embedding $u=g(x)$ and a stored vector $v$, similarity is computed as $\operatorname{cos}(u,v)=u^{\top}v$. This configuration restricts retrieval to distributionally aligned evidence within each body-location shard, thereby enabling robust local grounding without parameter updates.

\paragraph{Class-Wise Multi-Sensor Retrieval.}
Given a query window $x$ with sensor placement tag \textit{loc} and candidate activities $\mathcal{A}=\{a_1,\dots ,a_M\}$, we first compute its normalized embedding $u=g(x)$. We then perform \textit{class-conditional retrieval} by scoring $u$ against all historical vectors $v \in \mathcal{D}^{\text{loc}}$ labelled as $a_m$, producing a similarity-sorted list $\mathcal{L}_{m}^{\text{loc}}$. In multi-location sensor scenarios (e.g., wrist and ankle), we perform retrieval independently for each placement and fuse the rankings via Reciprocal Rank Fusion (RRF)~\cite{10.1145/1571941.1572114}:
\begin{equation*}
  \operatorname{RRF}(d)=
\sum_{\text{loc}}
\frac{1}{k_{\text{rrf}}+r_{\text{loc}}(d)},
\quad
k_{\text{rrf}}=60
\end{equation*}
where $r_{\text{loc}}(d)$ is the rank of document index $d$. Since indices are time-synchronized across sensors, this summation naturally aligns and jointly reranks multi-sensor evidence, promoting time windows that are consistently salient across modalities. Furthermore, as retrieval is performed conditionally per class, the agent receives the best available evidence for every hypothesis, ensuring balanced recall even for long-tail activities often overshadowed in global retrieval.

\paragraph{Hierarchical Multi-Agent Reasoning.}
ZARA orchestrates a hierarchical reasoning pipeline comprising three specialized agent roles executed in four stages (Figure~\ref{fig:prompt_flow}). Initially, a \emph{Feature Selector} agent queries the pairwise knowledge base $\mathcal{K}$ relative to the global candidate set $\mathcal{A}$ to identify $n$ coarse-grained discriminative features. Subsequently, an \emph{Evidence Pruning} agent synthesizes the retrieved class-wise evidence lists $\{\mathcal{N}_{a}(x)\}$ into a structured statistical comparison table (contrasting query values against class moments) based on these features. It then filters out distributionally mismatched activities, yielding a refined set $\mathcal{A}'$. The \emph{Feature Selector} is then re-engaged on $\mathcal{A}'$ to retrieve $m$ fine-grained features, enabling the system to resolve subtle ambiguities among the remaining candidates. Finally, a \emph{Decision Insight} agent analyzes the updated statistics to derive the final label $a'$, producing a transparent natural-language rationale grounded in the selected statistical features and retrieved evidence. Appendix~\ref{apd_prompt} provides the full prompts for each agent.

\begin{table*}[t]
    \centering
    \small
    \resizebox{\textwidth}{!}{
        \begin{tabular}{l |
        >{\columncolor{gray!10}}c>{\columncolor{gray!10}}c cc >{\columncolor{gray!10}}c>{\columncolor{gray!10}}c 
        cc >{\columncolor{gray!10}}c>{\columncolor{gray!10}}c cc 
        >{\columncolor{gray!10}}c>{\columncolor{gray!10}}c cc |
        >{\columncolor{gray!20}}c>{\columncolor{gray!20}}c}
        \toprule
        \multirow{4}{*}{\textbf{Dataset}} 
        & \multicolumn{6}{c }{\textbf{Easy}} & \multicolumn{6}{c }{\textbf{Medium}} & \multicolumn{4}{c|}{\textbf{Hard}} & \multicolumn{2}{c}{}  \\
        \cmidrule[0.9pt](lr){2-7} \cmidrule[0.9pt](lr){8-13} \cmidrule[0.9pt](lr){14-17} \cmidrule[0.9pt](lr){18-19}
        
        &\multicolumn{2}{>{\columncolor{gray!10}}c}{\textbf{Opportunity}}  &\multicolumn{2}{c}{\textbf{UCI-HAR}}  &\multicolumn{2}{>{\columncolor{gray!10}}c}{\textbf{Shoaib}} 
        &\multicolumn{2}{c}{\textbf{PAMAP2}} &\multicolumn{2}{>{\columncolor{gray!10}}c}{\textbf{USC-HAD}} &\multicolumn{2}{c}{\textbf{MHealth}} 
        &\multicolumn{2}{>{\columncolor{gray!10}}c}{\textbf{WISDM}} &\multicolumn{2}{c|}{\textbf{DSADS}} 
        &\multicolumn{2}{>{\columncolor{gray!20}}c}{\textbf{Average}} \\

        & \textbf{Acc} & \textbf{F1}  & \textbf{Acc} & \textbf{F1}  & \textbf{Acc} & \textbf{F1}  & \textbf{Acc} & \textbf{F1}  & \textbf{Acc} & \textbf{F1}  & \textbf{Acc} & \textbf{F1}  & \textbf{Acc} & \textbf{F1}  & \textbf{Acc} & \textbf{F1}  & \textbf{Acc} & \textbf{F1}  \\
        \midrule
        \# Classes  &\multicolumn{2}{>{\columncolor{gray!10}}c}{4} &\multicolumn{2}{c}{6} &\multicolumn{2}{>{\columncolor{gray!10}}c}{7} 
        &\multicolumn{2}{c}{12} &\multicolumn{2}{>{\columncolor{gray!10}}c}{12} &\multicolumn{2}{c }{12} 
        &\multicolumn{2}{>{\columncolor{gray!10}}c}{18} &\multicolumn{2}{c|}{19} 
        &\multicolumn{2}{>{\columncolor{gray!20}}c}{} \\
        \midrule
        \# Channels  &\multicolumn{2}{>{\columncolor{gray!10}}c}{30} &\multicolumn{2}{c}{6} &\multicolumn{2}{>{\columncolor{gray!10}}c}{30} 
        &\multicolumn{2}{c}{18} &\multicolumn{2}{>{\columncolor{gray!10}}c}{6} &\multicolumn{2}{c}{15}
        &\multicolumn{2}{>{\columncolor{gray!10}}c}{6} &\multicolumn{2}{c|}{30} &\multicolumn{2}{>{\columncolor{gray!20}}c}{} \\
        \midrule \midrule
        $\text{HARGPT}_{\text{ Text}}$ & 21.0 & 19.2 & 29.6 & 17.4  & 27.1  & 19.2 & 12.1  & 6.2  &13.8  &7.3 & 12.1 & 6.0 & 5.6 & 1.8 & 10.5 & 6.2 & 16.5 & 10.4 \\

        $\text{Gemini}_{\text{ Text}}$ & 26.5 & 19.8 & 24.2 & 13.0 & 27.1 & 17.6  & 15.0 & 10.2 & 14.2 & 5.4 & 25.4 & 20.8 & 11.1 & 7.6 & 13.2 & 8.6 & 19.6 & 12.9 \\ 

        $\text{Gemini}_{\text{ Table}}$ & 29.0 & 22.3  & 21.3  & 9.8  & 27.6  & 18.7  & 11.7   & 7.2 & 17.1 & 9.7  & 22.9  & 18.3 & 10.1  & 7.8  & 16.3 & 10.3 & 19.5  & 13.0 \\
        \midrule
        
        $\text{HARGPT}_{\text{ Plot}}$ & 21.5 & 15.6  & 28.3 & 15.7 & 24.3 & 14.2 & 10.0   & 6.9 & 14.6 & 8.7  & 15.0   & 11.0  & 5.9 & 2.8 & 7.9  & 4.5 & 15.9 & 9.9  \\

        $\text{Gemini}_{\text{ Plot}}$ & 23.5  & 21.3  & 20.6  & 31.7 & 31.4 & 24.1 & 10.4 & 6.9 & 10.8 & 5.3 & 19.2 & 17.4 & 9.4 & 7.2 & 10.0 & 4.8 & 18.3 & 13.5  \\
         \midrule
         
        NormWear & 23.0 & 23.8 & 17.9 & 11.4 & 15.2 & 11.7 & 9.2 & 2.7 & 10.0 & 5.8 & 8.3 & 2.2 & 4.2 & 1.4 & 3.7 & 2.2 & 11.4 & 7.7 \\

        IMUGPT & 38.5 & 28.7 & 32.5 & 21.6 & 26.7 & 15.2 & 12.9 & 3.8 & 2.9 & 1.9 & 8.3 & 2.8 & 5.9 & 2.1 & 7.4 & 3.6 & 16.9 & 10.0\\
                
        ImageBind & 35.5  & 30.0 & 28.8 & 19.9 & 36.7 & 30.2  & 18.8  & 10.2 & 7.9 & 1.8 & 17.9 & 11.1 & 8.0 & 4.7 & 10.5 & 5.7 & 20.5 & 14.2 \\

        IMU2CLIP & 36.5 & 34.4 & 33.3 & 22.8 & 39.5 & 34.5  & 15.8 & 11.6 & 16.3 & 10.5 & 16.3 & 14.3 & 10.1 & 5.9 & 13.7 & 9.2 & 22.7 & 17.9\\

        UniMTS &33.5 &24.8 &37.1  &23.9 &51.9 &40.6   &32.9   &29.2 &29.6   &24.2  &65.4 &58.8   &30.2   &28.5 & 34.7  &27.0 & 39.4  & 32.1  \\
        \midrule \midrule

        $\textbf{ZARA}_{\textbf{ Qwen-30B}}$ &84.0  & 84.2  &80.0  & 79.7  &91.9  & 91.7  &71.3  & 71.3  &42.1  & 41.4  &69.6  &69.0   &53.5  &50.9   &75.3 &73.5   &71.0  & 70.2  \\

        $\textbf{ZARA}_{\textbf{ Qwen-80B}}$ &80.5  &  79.8 &76.7  & 75.4  &\underline{93.8}  & \underline{93.5}  &67.5  & 67.0  & 47.1 & 49.2  &82.1  & \underline{81.6}  &57.6  &56.4   &81.6  &81.3  & 73.4  & 73.0  \\

         $\textbf{ZARA}_{\textbf{ GPT}}$ &\underline{86.5}  & \underline{86.5}  &\underline{85.0}  &  \underline{85.0}  &93.3  & 93.2  &\underline{72.5}  &\underline{72.8}   &\underline{56.7}  &  \underline{57.4} &\underline{82.5 } &80.6   &\underline{60.8}  & \underline{59.5}  & \underline{82.6 }&\underline{82.3}  & \underline{77.5} &  \underline{77.2} \\
        
        $\textbf{ZARA}_{\textbf{ Gemini}}$ &\textbf{92.5} &\textbf{92.5} &\textbf{90.0} &\textbf{90.0}  & \textbf{97.1} &\textbf{97.1}   & \textbf{76.7}    & \textbf{76.9}  &\textbf{60.0}   &\textbf{60.1}  &\textbf{86.3}    &\textbf{86.1}  &\textbf{65.6}    &\textbf{64.1}  & \textbf{84.2}   & \textbf{84.4} & \textbf{81.6} & \textbf{81.4}\\
        \bottomrule
        \end{tabular}
    }
    \caption{Cross-Subject Evaluation: ZARA vs. 10 baselines from three method families. Best scores are shown in \textbf{bold}; second-best are \underline{underlined}.}
    \label{tab:zeroshot}
% \vspace{-0.5em}
\end{table*}

\section{Experiments}

We adopt a two-tier evaluation protocol to assess ZARA's generalization: (1) \textit{Cross-Subject Generalization}, which tests robustness to individual variation within a domain; and (2) \textit{Cross-Dataset Generalization} (Section~\ref{exp:cross_dataset}), which evaluates transfer across heterogeneous sensor hardware and environments.

\subsection{Datasets}
We benchmark ZARA on 8 open-source and anonymized HAR datasets, ensuring ethical compliance and reproducibility. We group them into three levels: \textit{Easy} (Opportunity~\cite{Roggen2010CollectingCA}, UCI-HAR~\cite{Anguita2013APD}, Shoaib~\cite{s140610146}), \textit{Medium} (PAMAP2~\cite{6246152}, USC-HAD~\cite{10.1145/2370216.2370438}, MHealth~\cite{Baos2014mHealthDroidAN}), and \textit{Hard} (WISDM~\cite{wisdm_smartphone_and_smartwatch_activity_and_biometrics_dataset__507}, DSADS~\cite{ALTUN20103605}). 

\begin{figure}[t]
\centering
\includegraphics[width=1\linewidth]{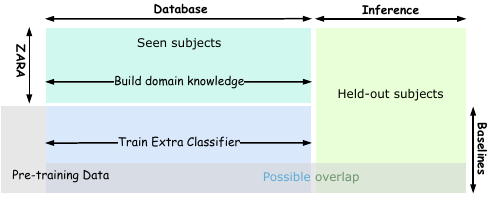} 
\caption{Subject split and data flow. 
ZARA builds its vector database and domain knowledge from seen subjects and tests on held-out subjects. 
Baselines may load pretrained weights or train a classifier head on seen subjects before testing on the same held-out subjects.
}
\label{fig:datasplit}
% \vspace{-0.5em}
\end{figure}

\subsection{Cross-Subject Generalization} \label{exp:cross_subject}

\paragraph{Setup.}
We evaluate ZARA's robustness using both open-source and proprietary LLMs: Qwen-3 (30B~\footnote{qwen3-30b-a3b-instruct-2507} and 80B~\footnote{qwen3-next-80b-a3b-instruct})~\cite{qwen3}, GPT-4.1-mini~\cite{openai2024gpt4technicalreport}, and Gemini-2.0-Flash~\cite{geminiteam2025geminifamilyhighlycapable}, with all agents set to temperature 0 for deterministic reproducibility. We employ a rigorous Subject-Hold-Out protocol (Figure~\ref{fig:datasplit}), where the knowledge base and retrieval index are distilled exclusively from \textit{Seen Subjects}, while inference is performed on a class-balanced split of \textit{Held-out Subjects}. This simulates a realistic deployment where the system must adapt to new users without calibration or fine-tuning. Additionally, to showcase scalability in large-scale settings, we replace static candidate lists with dynamic retrieval for the larger WISDM and DSADS benchmarks. For each query, we calculate the cosine similarity against the vector database to dynamically select the top-10 most relevant classes, thereby decoupling inference cost from the registered activity library size while preserving high recall. Appendix~\ref{apd_cross_subj} provides sensors, activities, statistics, and preprocessing details for each benchmark.

\begin{figure*}[t]
    \centering
    \begin{minipage}[c]{0.7\textwidth}
        \centering
        \includegraphics[width=\linewidth]{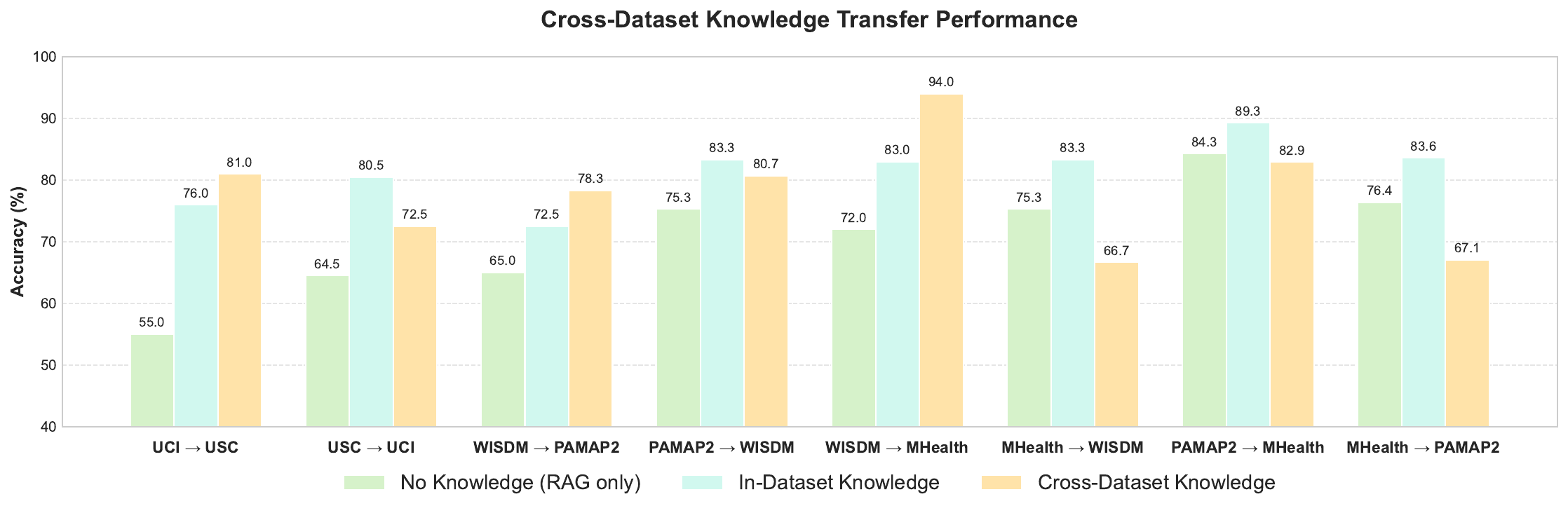}
    \end{minipage}
    \hfill 
    \begin{minipage}[c]{0.28\textwidth}
        \centering
        \resizebox{\linewidth}{!}{
            \begin{tabular}{l c l}
                \toprule
                \textbf{Cross-Dataset Pair} & \textbf{\# Act.} & \textbf{Sensor Position Mapping} \\
                \midrule
                UCI $\leftrightarrow$ USC & 5  & Waist $\leftrightarrow$ Front right hip \\
                \midrule
    
                PAMAP2 $\leftrightarrow$ WISDM & 5  & Wrist $\leftrightarrow$ Hand \\
                \midrule

                MHealth $\leftrightarrow$ WISDM & 5  & Right wrist $\leftrightarrow$ Hand \\
                \midrule
                
                \multirow{3}{*}{PAMAP2 $\leftrightarrow$ MHealth} & \multirow{3}{*}{7} & Wrist $\leftrightarrow$ Right wrist \\
                 &  & Chest $\leftrightarrow$ Chest \\
                 &  & Ankle $\leftrightarrow$ Left ankle\\

                \bottomrule
            \end{tabular}
        }
    \end{minipage}
    \caption{Cross-Dataset Evaluation. \textbf{Left:} Accuracy under No-Knowledge, In-Dataset Knowledge, and Cross-Dataset Knowledge settings. \textbf{Right:} dataset compatibility in terms of shared activities and sensor position mapping.}
    \label{fig:cross_dataset_main}
    % \vspace{-0.5em}
\end{figure*}

\paragraph{Baselines.}
To ensure a rigorous and fair comparison in training-free settings, we benchmark ZARA against 10 representative baselines that align with the data flow in Figure~\ref{fig:datasplit}. These models rely on either pre-training on external datasets or implicit LLM knowledge to perform inference on held-out subjects. We categorize them into three families: 
(i) \textit{Text-based LLMs}: $\text{HARGPT}_{\text{Text}}$~\cite{ji2024hargptllmszeroshothuman}, $\text{Gemini}_{\text{Text}}$ and $\text{Gemini}_{\text{Table}}$. The latter follows the structured input protocol described in~\cite{fang2024large, 10.1145/3746252.3761056}, where time-series data is encoded as a Markdown table;
(ii) \textit{Multimodal LLMs}: $\text{HARGPT}_{\text{Plot}}$ and $\text{Gemini}_{\text{Plot}}$, which leverage plotted sensor signals as visual prompts for activity prediction; 
(iii) \textit{Pretrained HAR Models}: ImageBind~\cite{girdhar2023imagebind}, IMU2CLIP~\cite{moon-etal-2023-imu2clip}, NormWear~\cite{luo2024foundationmodelmultivariatewearable}, and UniMTS~\cite{zhang2024unimtsunifiedpretrainingmotion}, which learn modality-aligned embeddings for training-free transfer. We also include IMUGPT~\cite{Leng2023generating}, which uniquely differs by pretraining on task-specific virtual motion data for downstream deployment. Further details are provided in Appendix~\ref{apd_baselines}.

\paragraph{Results.}
Table~\ref{tab:zeroshot} reports cross-subject performance in parameter-frozen settings. ZARA consistently outperforms all baselines across all difficulty levels. Our best variant, $\text{ZARA}_{\text{Gemini}}$, achieves an average accuracy of 81.6\% and a macro F1 of 81.4\%, substantially surpassing the strongest baseline, UniMTS. This advantage holds across backbone scales: ZARA variants powered by Qwen3-30B/80B and GPT-4.1-mini all outperform every baseline, indicating that the gains arise from the robustness of ZARA's knowledge- and retrieval-augmented agentic framework rather than backbone size. Performance differences across backbones are mainly attributable to their numerical reasoning ability, particularly when interpreting statistical distribution tables for inference.

In contrast, existing methods exhibit fundamental limitations. IMUGPT fails to transfer from virtual pre-training to real-world benchmarks, while contrastive approaches (ImageBind, IMU2CLIP) are restricted to single-sensor inputs. Even multi-sensor models such as UniMTS and NormWear degrade sharply on activities outside their pre-training distributions. A common failure mode across baselines is a large gap between accuracy and macro F1, revealing a strong bias toward majority classes under distribution shift. By contrast, ZARA maintains close alignment between accuracy and F1, demonstrating robust recognition of long-tail activities via class-balanced retrieval. Moreover, direct prompting methods (HARGPT, $\text{Gemini}_{\text{Text/Table/Plot}}$) fail catastrophically, highlighting that without explicit reference grounding, even capable LLMs cannot reason over numerical sensor streams. Overall, ZARA provides an interpretable, evidence-grounded solution that substantially improves the reliability of training-free inference for HAR. Detailed token-usage statistics for each agent stage are provided in Appendix~\ref{apd_token_usage}.

\subsection{Cross-Dataset Generalization} \label{exp:cross_dataset}

\paragraph{Setup.}
Moving beyond subject variations, this tier evaluates robustness against sensor heterogeneity through transfer experiments across distinct dataset pairs. As illustrated in Figure~\ref{fig:cross_dataset_main} (Right), we establish a \textit{Common Evaluation Protocol} restricted to the intersection of sensor placements and activity labels. All experiments in this section utilize Gemini-2.0-Flash as the backbone. Appendix~\ref{apd_cross_dataset} provides sensors, activities, and preprocessing details for each cross-dataset pair.

\begin{table*}[t]
    \centering
    \small
    \resizebox{\textwidth}{!}{
        \begin{tabular}{l |
        >{\columncolor{gray!10}}c>{\columncolor{gray!10}}c cc >{\columncolor{gray!10}}c>{\columncolor{gray!10}}c 
        cc >{\columncolor{gray!10}}c>{\columncolor{gray!10}}c cc 
        >{\columncolor{gray!10}}c>{\columncolor{gray!10}}c cc |
        >{\columncolor{gray!20}}c}
        
        \toprule
        \multirow{4}{*}{\textbf{Dataset}} 
        & \multicolumn{6}{c}{\textbf{Easy}} & \multicolumn{6}{c}{\textbf{Medium}} & \multicolumn{4}{c|}{\textbf{Hard}} & \multicolumn{1}{c|}{\textbf{}}  \\
        \cmidrule[0.9pt](lr){2-7} \cmidrule[0.9pt](lr){8-13} \cmidrule[0.9pt](lr){14-17}  \cmidrule[0.9pt](lr){18-18} 
        
        &\multicolumn{2}{>{\columncolor{gray!10}}c}{\textbf{Opportunity}}  &\multicolumn{2}{c}{\textbf{UCI-HAR}}  &\multicolumn{2}{>{\columncolor{gray!10}}c}{\textbf{Shoaib}} 
        &\multicolumn{2}{c}{\textbf{PAMAP2}} &\multicolumn{2}{>{\columncolor{gray!10}}c}{\textbf{USC-HAD}} &\multicolumn{2}{c}{\textbf{MHealth}} 
        &\multicolumn{2}{>{\columncolor{gray!10}}c}{\textbf{WISDM}} &\multicolumn{2}{c|}{\textbf{DSADS}} 
        &\textbf{Time (s)} \\

        & \textbf{Acc} & \textbf{F1}  & \textbf{Acc} & \textbf{F1}  & \textbf{Acc} & \textbf{F1}  & \textbf{Acc} & \textbf{F1}  & \textbf{Acc} & \textbf{F1}  & \textbf{Acc} & \textbf{F1}  & \textbf{Acc} & \textbf{F1}  & \textbf{Acc} & \textbf{F1}  & \textbf{Avg}   \\
        \midrule \midrule
        
        \rowcolor{gray!20}
        \multicolumn{18}{l}{\textit{Frozen Embedder + Supervised Head}} \\
        
        Moment-S   &66.0 &64.8  &77.5 &77.5  &86.2 &85.9    &71.7 &71.8     &54.2   & 52.8   &67.5    &66.8  &\underline{66.3}  &\underline{66.3}  &72.6    &72.3 &--\\

        Moment-L       &63.5  &62.7   &78.8 &78.6  & 91.0 & 90.8  &73.3   &73.4    &47.1   &45.2   &72.1    &72.2   &65.3    &65.6   &74.2   &73.7 &--\\

        Mantis              &90.0   &89.9  &\textbf{91.3} &\textbf{91.2}  &93.3    &92.9   & \textbf{84.6} &\textbf{85.1}    &53.8    &53.7  &86.7     &86.0    &\textbf{71.5}   & \textbf{71.2}  &\textbf{90.5}  &\textbf{90.2} &-- \\
        \midrule 

        \rowcolor{gray!20}
        \multicolumn{18}{l}{\textit{Classifier-Free, Knowledge-Augmented Reasoning}} \\

        $\textbf{ZARA}_{\text{DTW}}$   &90.5  &90.5  &\underline{90.4}  &\underline{90.3}  &96.7    &96.7     &71.7  &71.6     &55.4     &56.4   & \underline{86.3}   & 86.1 &59.4  & 57.3  & 82.6  &82.6 &0.3826\\

        $\textbf{ZARA}_{\text{Moment-S}}$     &88.5   &88.4   &87.9 &87.7  &\textbf{97.6}   &\textbf{97.6}   &73.3   &73.4    &53.3   &53.0   &\underline{86.3}    & \underline{86.2} & 62.2  & 62.1  & \underline{86.3}  &\underline{86.0} &\textbf{0.0438}\\

        $\textbf{ZARA}_{\text{Moment-L}}$      &\underline{91.0}   &\underline{91.0}   &87.9   &87.8   &\textbf{97.6}   &\textbf{97.6}   &75.8    &76.1   &\underline{55.8}    &\underline{56.7}  &\textbf{88.3}    &\textbf{88.0}  & 65.3   & 64.2 &84.7   & 83.9 &\underline{0.1003}  \\

        $\textbf{ZARA}_{\text{Mantis}}$    &\textbf{92.5}   &\textbf{92.5}   &90.0  &90.0  &\underline{97.1}   &\underline{97.1}   & \underline{76.7}   & \underline{76.9}   &\textbf{60.0}   &\textbf{60.1}   &\underline{86.3}    &86.1  &65.6   &64.1  & 84.2 & 84.4  &0.1826  \\
        \bottomrule
        \end{tabular}
    }
    \caption{Ablation of Retrieval Backbones. Comparison between ZARA and supervised baselines across diverse representations (DTW, Mantis, Moment). The rightmost column indicates the average retrieval time per query.}
    \label{tab:retrieval}
    % \vspace{-0.5em}
\end{table*}

\paragraph{Baselines.}
To strictly isolate the contribution of transferable prior knowledge, we compare three internal settings. Crucially, all settings share the same retrieval backbone and agent workflow, varying only in the source of knowledge used to guide feature selection: (1) \textit{No Knowledge}, a baseline where the agent is provided with the full list of feature names and definitions, forcing the LLM to rely solely on its internal parametric knowledge to select discriminative features; (2) \textit{In-Dataset Knowledge}, an upper-bound setting utilizing a knowledge base constructed directly from the target dataset, representing the ideal scenario with perfect domain adaptation; and (3) \textit{Cross-Dataset Knowledge}, the proposed setting where the agent is guided by a knowledge base derived from a distinct source dataset, explicitly testing the transferability of motion priors.

\paragraph{Results.}
Figure~\ref{fig:cross_dataset_main} (Left) reports transfer performance across 8 scenarios. To mitigate domain shifts arising from differences in sensor hardware across datasets, we restrict transfer to the \textit{motion priors} derived from the source. This design mirrors practical deployments, where wearable devices often differ across users, brands, and generations, making it impractical to reuse raw-signal evidence collected under a different sensor configuration. Conversely, the retrieval database is constructed from the target domain, ensuring that the \textit{statistical feature evidence} analyzed by the LLM remains distribution-aligned with the query. This setup reveals a clear relationship between knowledge quality and transfer performance. While in-dataset knowledge performs best in 5 cases, cross-dataset knowledge unexpectedly outperforms it in 3 transfers: UCI$\to$USC, WISDM$\to$PAMAP2, and WISDM$\to$MHealth. 

This asymmetry is driven by two key factors. First, \textit{User Diversity} strongly influences transferability: knowledge transferred from highly diverse sources such as WISDM (39 subjects) to lower-diversity targets like PAMAP2 (7 subjects) and MHealth (8 subjects) yields substantial gains, suggesting that knowledge derived from diverse populations captures more transferable motion priors that generalize better than local knowledge overfitted to a small cohort. Second, \textit{Data Density} plays a critical role. Transfers from data-rich datasets such as PAMAP2 ($\sim$4.3k samples) to data-scarce targets like MHealth ($\sim$1.7k samples) retain higher performance than the reverse direction, indicating that dense data distributions are essential for learning fine-grained motion priors. Notably, when source knowledge quality is low (e.g., MHealth as the source), cross-dataset performance matches or falls below the No-Knowledge baseline. This behavior suggests that the agent functions as a grounded reasoning engine: it leverages injected knowledge when informative, but does not hallucinate gains beyond the quality of the provided knowledge.

\section{Ablation Studies}
\label{ablation}

In this section, we ablate key components of ZARA to quantify their individual impact. We fix Gemini-2.0-Flash as the backbone. Appendix~\ref{apd_ablation} provides detailed results for each ablation.

\paragraph{Impact of Retrieval Embedder.}
To evaluate the contribution of our retrieval strategies, we compare ZARA with non-LLM retrieval and supervised baselines built on different representations under four backbone settings (see details in Appendix~\ref{apd_retrive_backbone}): classical DTW~\cite{Müller2007} and three pre-trained foundation models, Moment-Small/Large~\cite{goswami2024moment} and Mantis~\cite{feofanov2025mantislightweightcalibratedfoundation}. Moment is pre-trained via masked TS prediction, while Mantis is optimized for TS classification and explicitly incorporates HAR datasets during pre-training. We adhere to strict evaluation protocols: baselines train supervised classifiers on frozen embeddings, whereas ZARA utilizes these embeddings strictly as anchors for parameter-frozen retrieval. 
As shown in Table~\ref{tab:retrieval}, ZARA exhibits strong robustness across embedders, maintaining high accuracy in parameter-frozen settings regardless of backbone choice. Notably, ZARA frequently outperforms its supervised counterparts despite lacking task-specific parameter optimization. With Moment-Small, ZARA surpasses the baseline on 7/8 datasets in F1; with Moment-Large, it wins 7/8 datasets on both metrics; and with Mantis, it improves F1 on 4/8 datasets. Overall, ZARA ranks first on 4 datasets and second on 7.
These results confirm that ZARA's knowledge- and retrieval-augmented agentic pipeline delivers substantial reasoning gains beyond raw embedding similarity, enabling classifier-free generalization without the training overhead required by supervised baselines.

Latency profiling (Table~\ref{tab:retrieval}) reports the average time required to process a single query on an Apple M2 Max CPU with 64GB memory. Moment-Small is the fastest, whereas DTW is substantially slower due to costly pairwise sequence alignment. Although Mantis is lightweight, it incurs higher latency than Moment because it concatenates channel-wise embeddings rather than averaging them; in return, it retrieves more informative samples, reflecting a speed–quality trade-off. While absolute latency varies with query data, database size, and hardware, the relative rankings consistently capture method-level efficiency.

\paragraph{Removing Retrieval Reduces Performance.}

\begin{figure}[t]
\centering
\includegraphics[width=1\linewidth]{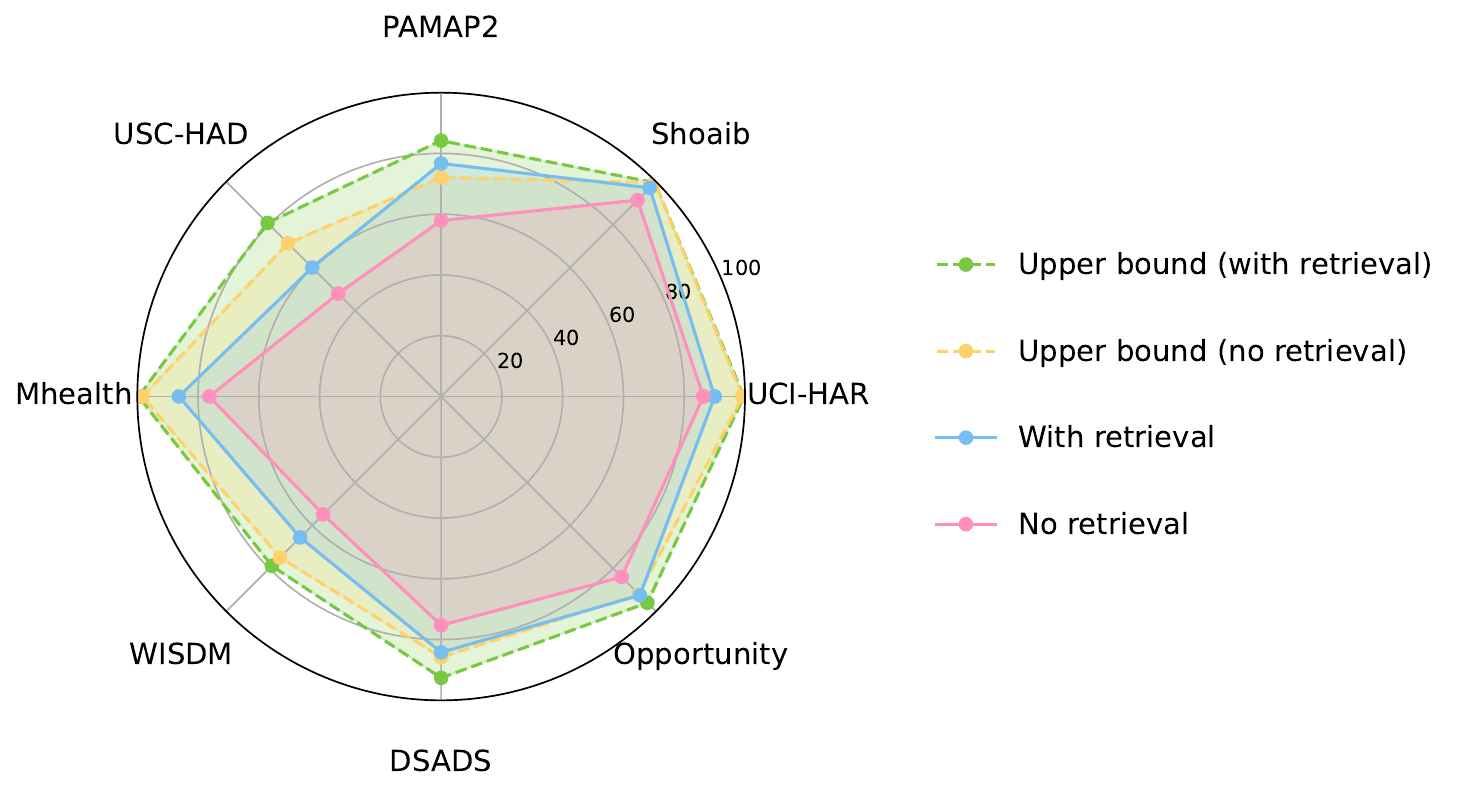} 
\caption{Impact of retrieval. The \textit{upper bound} denotes the proportion of queries for which the pruned candidate set retains the ground-truth label under each setting.}
\label{fig:retrieval_ablation}
\end{figure}

To assess the necessity of local evidence anchoring, we ablate the retrieval module by replacing top-k retrieval with global class-wise feature distributions computed over the entire database. This forces the LLM to reason solely on global priors without instance-level grounding. As shown in Figure~\ref{fig:retrieval_ablation}, this degradation significantly hampers ZARA's training-free reasoning: average accuracy falls from 81.6\% to 71.8\%, and the upper bound (the retention rate of the ground-truth label) drops from 91.4\% to 86.7\%. The decline is particularly pronounced in datasets where individual instance statistics diverge from global averages. These results confirm that parameter-frozen inference requires more than just abstract knowledge; retrieval is essential to surface query-relevant evidence that bridges the gap between global statistics and local signal dynamics.

\paragraph{Skipping Evidence Pruning Hurts.}

\begin{figure}[t]
\centering
\includegraphics[width=1\linewidth]{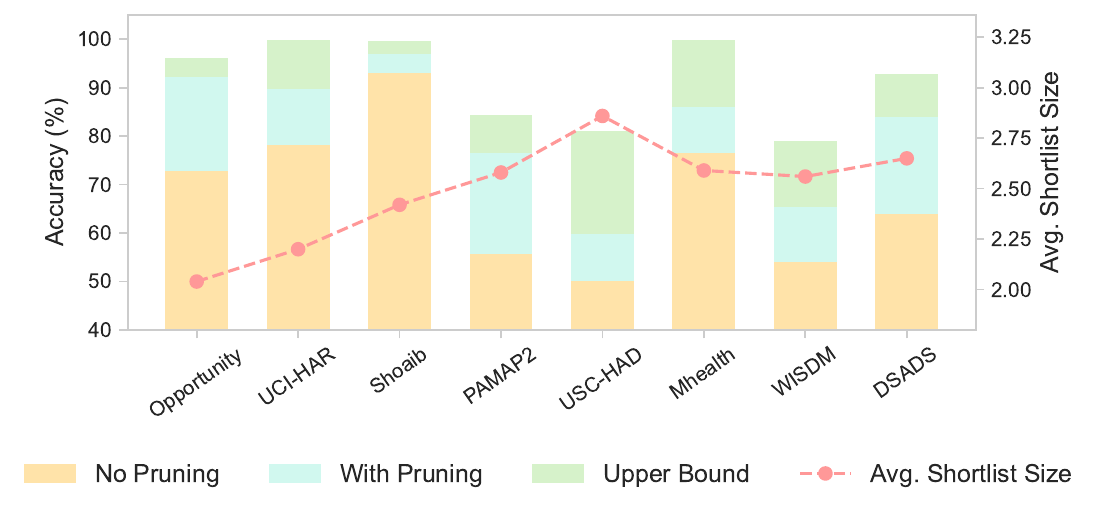} 
\caption{Accuracy with and without the \textit{Evidence Pruning} Agent, along with upper bounds for each setting. The dashed line indicates the average length of the pruned candidate set.}
\label{fig:prune}
% \vspace{-0.5em}
\end{figure}

To quantify the role of the Evidence Pruning Agent, we ablate it across all eight benchmarks. Removing pruning causes a sharp drop in average accuracy in parameter-frozen settings, from 81.6\% to 68.2\%. Figure~\ref{fig:prune} shows that our pruning agent typically narrows each query to 2--3 candidates per benchmark, with the easy-level datasets yielding even smaller shortlists. Moreover, this aggressive reduction is achieved with minimal information loss, maintaining a 91.4\% upper-bound accuracy (ground-truth retention rate). By filtering out clearly mismatched activities early, the system allows the LLM to focus its reasoning capacity on distinguishing the remaining hard negatives using finer-grained evidence. In contrast, omitting pruning forces the LLM to reason over the full candidate pool, degrading focus and performance. Notably, even in the ablated setting, ZARA’s 68.2\% still outperforms every baseline.

\paragraph{No Prior Knowledge Fails.}

\begin{figure}[t]
\centering
\includegraphics[width=1\linewidth]{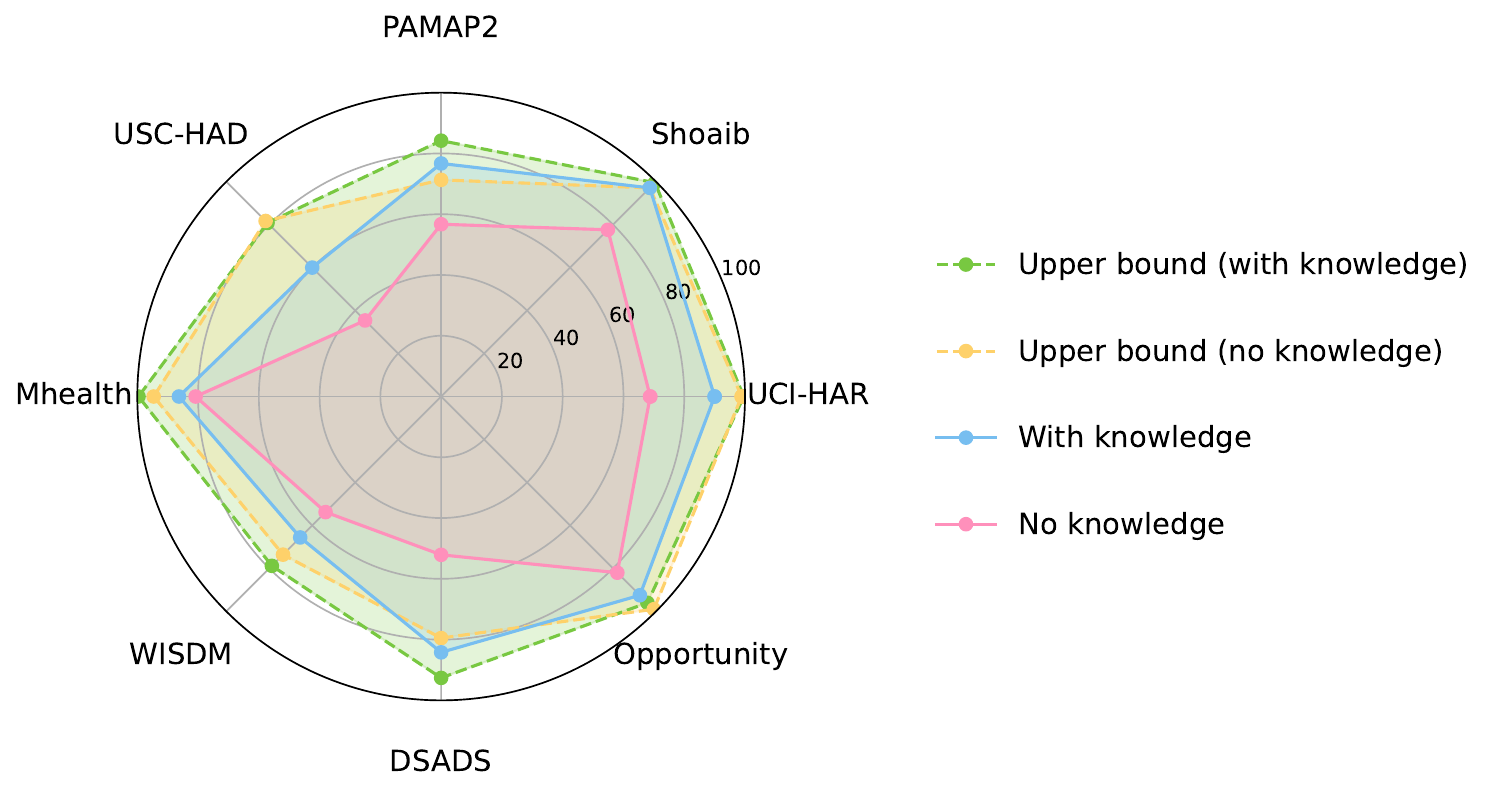} 
\caption{Impact of the prior knowledge base. The \textit{upper bound} denotes the proportion of queries for which the pruned candidate set retains the ground-truth label under each setting.}
\label{fig:knowledge_ablation}
\end{figure}

To quantify the value of injected priors, we disable the pairwise feature-importance knowledge registry, forcing the Feature Selector to rely solely on the LLM's intrinsic world knowledge. As shown in Figure~\ref{fig:knowledge_ablation}, this ablation causes a sharp decline in average accuracy in parameter-frozen settings, from 81.6\% to 63.4\%. The average upper-bound (ground-truth retention) also drops from 91.4\% to 87.0\%, indicating that without statistical grounding, the agent struggles to identify discriminative motion properties. Benchmarks with fewer classes (e.g., Opportunity, UCI-HAR, Shoaib) are less affected in the first narrowing stage, which mainly removes clearly irrelevant classes. In the second stage, however, the Feature Selector often chooses suboptimal features due to the absence of effective criteria, leading to overly coarse feature selection that cannot recover the lost accuracy. This confirms that general-purpose LLMs cannot reliably infer discriminative motion priors from text alone; they require statistically grounded references for robust training-free inference.

\section{Conclusions}
\label{conclusion}

We present ZARA, an agentic framework that combines statistical profiling, retrieval augmentation, and hierarchical reasoning to translate implicit sensor dynamics into explicit linguistic priors. Rather than treating labeled data solely as training targets, ZARA reinterprets them as a reference registry for grounded reasoning. This design anchors LLM decision-making in real-world evidence, enabling training-free inference that is verifiable rather than purely generative, while allowing off-the-shelf LLMs to be deployed without task-specific adaptation.

Extensive experiments across eight HAR benchmarks demonstrate that ZARA achieves state-of-the-art performance in both cross-subject and cross-dataset settings. The results show that ZARA captures transferable motion priors that generalize across diverse user populations and heterogeneous sensor domains. By producing transparent predictions grounded in retrieved evidence, ZARA offers a scalable and trustworthy path toward adaptive HAR in the wild.

\section{Limitations}
\label{limitation}

Direct architectural comparisons remain limited, as ZARA occupies a relatively distinct design space at the intersection of domain knowledge, retrieval augmentation, and agentic reasoning for HAR. To ensure a rigorous evaluation, we compare against strong foundation-model baselines (e.g., UniMTS, ImageBind, Mantis, and Moment), whose extensive \textit{domain pre-training} or \textit{task-specific classifiers} serve as practical alternatives to ZARA's retrieval-grounded inference pipeline. We do not include recent methods such as SensorLM~\cite{zhang2025sensorlmlearninglanguagewearable}, LLaSA~\cite{imran2025llasasensorawarellmnatural}, and RelCon~\cite{xu2025relcon} because of fundamental input-modality mismatches. While ZARA is designed for synchronized, multi-location motion-sensor data, these methods are typically restricted to single-location inputs, unimodal accelerometer streams, or non-motion modalities (e.g., temperature), making direct comparison infeasible on our standardized multi-view motion benchmarks.

In addition, although ZARA is parameter-frozen and avoids task-specific optimization at inference time, it still requires a labeled retrieval database to ground reasoning in representative evidence. This support set is not used to update model weights, but functions as a reference registry that links implicit sensor patterns to explicit, verifiable comparisons. Finally, the agentic workflow introduces higher computational cost than simpler unimodal pipelines. Future work will therefore explore more efficient agentic designs and domain-adapted LLMs that reduce dependence on external references while preserving transparency and grounded reasoning.
\section{Acknowledgements}
\label{acknowledgements}

This work was supported by the ARC Centre of Excellence for Automated Decision-Making and Society (CE200100005).


\bibliography{custom}
\clearpage
\pagebreak

\appendix

\section{Appendix}
\label{appendix}
This appendix provides additional implementation and evaluation details supporting the main findings of the paper. We first describe the baseline models used for comparison, followed by dataset and preprocessing details. We then report per-dataset token usage for each agent and per-class performance across all benchmarks, complementing the aggregate results in the main text. Finally, we include the full prompts used at each reasoning stage in ZARA.

\subsection{Baselines}
\label{apd_baselines}
We provide implementation details for all baselines used in our study, including how each method was reproduced or adapted for evaluation in our training-free HAR setting.

\paragraph{HARGPT~\cite{ji2024hargptllmszeroshothuman}.}
This method directly prompts LLMs to classify motion time-series. We follow the original setup by downsampling input signals to 10\,Hz and applying the prompt template for raw numerical input. We additionally evaluate a visual-input variant by providing plotted sensor signals to the underlying LLM (GPT-4o-mini~\cite{openai2024gpt4o}). For visual inputs, each 6-channel sensor is rendered as an individual subplot, and multiple sensors are concatenated into a single composite figure to reduce visual clutter in multi-sensor datasets.

\paragraph{Gemini~\cite{geminiteam2025geminifamilyhighlycapable}.}
To assess the improvements brought by ZARA, we use Gemini-2.0-Flash, the same LLM backbone adopted in our framework, as a standalone baseline. Similar to HARGPT, Gemini is evaluated with raw numerical sequences and plotted sensor signals. In addition, we include a third input modality: Markdown-formatted tables, to test whether more structured input alone improves recognition accuracy. These Gemini baselines allow us to isolate the contribution of ZARA's retrieval, knowledge, and agentic reasoning modules beyond the capability of the base model itself.

\paragraph{ImageBind~\cite{girdhar2023imagebind}.}
ImageBind learns a unified embedding space across six modalities: image, text, audio, depth, thermal, and IMU. We use the publicly released \texttt{imagebind\_huge} checkpoint for evaluation in our training-free setting. Since ImageBind only supports single-sensor input and requires fixed-length windows of size $6\times 2000$, we evaluate each sensor placement separately. To satisfy the input-length requirement, we apply two strategies---repeat padding and linear interpolation to 2000 steps---and report the best result across sensor placements for each dataset.

\paragraph{IMU2CLIP~\cite{moon-etal-2023-imu2clip}.}
IMU2CLIP aligns inertial measurement unit (IMU) data with video and text by projecting them into CLIP's joint embedding space. Similar to ImageBind, it only supports single-sensor input and requires fixed-length windows of size $6\times 1000$. We therefore evaluate each sensor placement separately and use both repeat padding and interpolation to match the required input length, reporting the best result for each dataset.

\paragraph{NormWear~\cite{luo2024foundationmodelmultivariatewearable}.}
NormWear is a foundation model designed to extract generalized representations from multivariate wearable signals. It is pre-trained on a diverse corpus of physiological data, including PPG, ECG, EEG, GSR, and IMU, collected from multiple public datasets. For activity recognition in our training-free setting, we follow the official documentation and use the recommended prompt, ``What is the activity being performed currently?'', together with the candidate activity labels for inference.

\paragraph{IMUGPT~\cite{Leng2023generating}.}
IMUGPT generates synthetic training data by first prompting GPT-4o-mini to produce diverse textual activity descriptions. These descriptions are converted into 3D motion sequences and then into virtual IMU streams. For evaluation, we adopt DeepConvLSTM, the best-performing backbone reported in the original paper. To ensure a fair comparison in our training-free setting, we exclude the supervised distribution calibration stage, which depends on labeled downstream data.

\paragraph{UniMTS~\cite{zhang2024unimtsunifiedpretrainingmotion}.}
UniMTS proposes a unified pretraining framework for motion time-series that generalizes across diverse device configurations, including sensor position and orientation. It uses contrastive learning to align motion signals with LLM-enriched text descriptions, enabling classifier-free recognition through semantic matching. For our experiments, we follow the official implementation and evaluate UniMTS in its training-free inference setting, using the released checkpoints and the corresponding text-label matching protocol.

\subsection{Retrieval Strategies}
\label{apd_retrive_backbone}

All retrieval strategies in ZARA follow a two-stage pipeline. First, candidates are ranked independently within each sensor placement according to their similarity to the query. The ranked lists are then aggregated across placements using Reciprocal Rank Fusion (RRF) to obtain the final retrieval results. We evaluate four retrieval backbones: DTW and three pretrained time-series encoders.

\paragraph{Dynamic Time Warping (DTW)~\cite{Müller2007}.}
We implement DTW using the multi-dimensional variant from the \texttt{dtaidistance} package~\cite{meert2021dtaidistance}. For each query segment and each database candidate, we first apply z-score normalization independently to each of the six sensor channels. We then compute the DTW distance between the query and each candidate segment using the \texttt{distance\_fast} routine with pruning enabled to accelerate matching. Distances are negated to obtain similarity scores, and the top-$k$ candidates are returned for retrieval.

\begin{table}[t]
\centering
\small
\resizebox{\linewidth}{!}{
\begin{tabular}{lcccc}
\toprule
\textbf{Dataset} & \multicolumn{2}{c}{\textbf{\# Subjects}} & \multicolumn{2}{c}{\textbf{\# Samples}} \\
\cmidrule(lr){2-3}\cmidrule(lr){4-5}
 & \textbf{Database} & \textbf{Inference} & \textbf{Database} & \textbf{Inference} \\
\midrule
Opportunity & 3  & 1 & 6968  & 200 \\
UCI-HAR     & 21 & 9 & 7352  & 240 \\
Shoaib      & 8  & 2 & 5040  & 210 \\
PAMAP2      & 7  & 2 & 7138  & 240 \\
USC-HAD     & 12 & 2 & 11889 & 240 \\
MHealth     & 8  & 2 & 2799  & 240 \\
WISDM       & 39 & 8 & 14287 & 288 \\
DSADS       & 6  & 2 & 6840  & 190 \\
\bottomrule
\end{tabular}}
\caption{Cross-subject dataset statistics.}
\label{tab:cross_subj_stats}
\end{table}

\paragraph{Moment~\cite{goswami2024moment}.}
Moment is a time-series foundation model based on the T5 architecture, pre-trained on a range of tasks including classification, anomaly detection, and forecasting. We evaluate both the \texttt{moment-small} and \texttt{moment-large} variants, whose embedding dimensions are 512 and 1024, respectively. For multi-channel inputs, Moment produces per-channel embeddings, which we average to obtain a single representation for retrieval in ZARA. In the supervised baseline comparison reported in Table~\ref{tab:retrieval}, we additionally follow the standard frozen-encoder protocol by training an SVM classifier on top of the database split and selecting hyperparameters through greedy search.

\paragraph{Mantis~\cite{feofanov2025mantislightweightcalibratedfoundation}.}
Mantis is a foundation model for time-series classification built on the Vision Transformer (ViT) architecture and pre-trained via contrastive learning. It has also been pre-trained on several HAR-related datasets. For input processing, Mantis rescales each time-series to a fixed length of 512, extracts a 256-dimensional embedding from each channel, and concatenates the channel-wise embeddings into a unified representation. We use this representation as the retrieval backbone in ZARA. In the supervised baseline comparison reported in Table~\ref{tab:retrieval}, we follow the frozen-encoder setting of the original method by training a random forest classifier on the database split.

\begin{table*}[t]
\centering
\small
\resizebox{\textwidth}{!}{
\begin{tabular}{lcll}
\toprule
\textbf{Dataset} & \textbf{\# Classes} & \textbf{Classes} & \textbf{Sensor Placements} \\ \midrule
Opportunity & 4 & 
Stand, Walk, Sit, Lie & 
 Back, upper arms, lower arms \\ \hline

 UCI-HAR & 6 & 
\begin{tabular}[c]{@{}l@{}}Standing, Sitting, Laying, Walking, Walking downstairs, \\Walking upstairs\end{tabular} & 
Waist \\ \hline

Shoaib & 7 & 
\begin{tabular}[c]{@{}l@{}}Walking, Standing, Jogging, Sitting, Biking, Downstairs,\\ Upstairs\end{tabular} & 
\begin{tabular}[c]{@{}l@{}}right pockets, left pockets, belt, \\ Right upper arm, right wrist\end{tabular} \\ \hline

PAMAP2 & 12 & 
\begin{tabular}[c]{@{}l@{}}Lying, Sitting, Standing, Ironing, Vacuum cleaning, \\ Ascending stairs, Descending stairs, Walking, \\ Nordic walking, Cycling, Running, Rope jumping\end{tabular} & 
Wrist, chest, ankle\\ \hline

USC-HAD & 12 & 
\begin{tabular}[c]{@{}l@{}}Sleeping, Sitting, Elevator down, Elevator up,\\ Standing, Jumping, Walking downstairs, Walking right, \\ Walking forward, Running forward, Walking upstairs, \\Walking left\end{tabular} & 
Front right hip \\ \hline

MHealth & 12 & 
\begin{tabular}[c]{@{}l@{}}Climbing stairs, Standing still, Sitting and relaxing, \\ Lying down, Walking, Waist bends forward, \\ Frontal elevation of arms, Knees bending (crouching), \\ Jogging, Running, Jump front \& back, Cycling\end{tabular} & 
Chest, right wrist, left ankle \\ \hline

WISDM & 18 & 
\begin{tabular}[c]{@{}l@{}}Walking, Jogging, Stairs, Sitting, Standing, Typing,\\ Brushing Teeth, Eating Soup, Eating Chips, Eating Pasta,\\ Eating Sandwich, Kicking Ball, Playing Catch Ball,\\ Drinking, Dribbling Ball, Writing, Clapping,\\Folding Clothes\end{tabular} & 
Hand \\ \hline

DSADS & 19 & 
\begin{tabular}[c]{@{}l@{}}Sitting, Standing, Lying on back, Lying on right side,\\Ascending stairs, Descending stairs, Standing in elevator,\\ Moving around in elevator, Walking slowly, Rowing, \\Jumping, Walking on a treadmill in flat positions, \\Walking on a treadmill in inclined positions,\\ Running on a treadmill fast, Exercising on a stepper,\\ Exercising on a cross trainer, Playing basketball,\\Cycling on an exercise bike in horizontal positions, \\ Cycling on an exercise bike in vertical positions\end{tabular} & 
\begin{tabular}[c]{@{}l@{}}Torso, right arm, left arm,\\right leg, left leg\end{tabular} \\ \bottomrule

\end{tabular}
}
\caption{Dataset classes and sensor placements.}
\label{tab:dataset_categories}
\end{table*}

\subsection{Data Preprocessing}
\label{apd_preprocess}

\subsubsection{Cross-Subject Generalization}
\label{apd_cross_subj}
This section introduces the data preprocessing pipeline for the Section~\ref{exp:cross_subject} Cross-Subject Generalization study. Due to the cost constraints of API-based inference and the need for detailed ablation analyses, we evaluate each dataset on a randomly sampled inference subset. For each dataset, we ensure balanced coverage by sampling an equal number of non-overlapping windows per activity class and per subject. This design strikes a practical balance between cost efficiency and diversity across datasets, activity types, and subjects. For fair comparison, we keep the sampled subsets identical across all baselines. Each dataset contains multiple activity classes, and the corresponding sensor placements are summarized in Table~\ref{tab:dataset_categories}. Table~\ref{tab:cross_subj_stats} reports the statistics for the database and test sets.

\paragraph{Opportunity~\cite{Roggen2010CollectingCA}.} The dataset contains recordings from 4 subjects at a sampling rate of 30\,Hz. We designate Subject 4 as the inference user and use data from the remaining subjects to build the retrieval database. Motion sensor data are segmented into non-overlapping 2-second windows (60 timesteps each). For Inference, we randomly sample 50 windows per activity class from the inference split, resulting in a balanced set of 200 samples.

\paragraph{UCI-HAR~\cite{Anguita2013APD}.} The dataset contains recordings from 30 volunteers, sampled at 50\,Hz. The dataset is pre-segmented using fixed-width sliding windows of 2.56 seconds with 50\% overlap. Following the original split, we use data from test set (9 subjects) for inference and the remaining for the database. From the inference set, we randomly sample 40 windows per activity, ensuring user-balanced representation within each class, resulting in a total of 240 samples.

\paragraph{Shoaib~\cite{s140610146}.} The dataset contains recordings from 10 subjects, sampled at 50\,Hz. We use data from subjects 1 and 9 for inference and the remaining subjects for the database. The recordings are segmented into non-overlapping windows of 2 seconds (100 timesteps). For inference, we randomly sample 30 windows per activity from the inference users (15 from each) yielding a class-balanced test set of 210 samples.

\paragraph{PAMAP2~\cite{6246152}.} This dataset contains recordings from 9 subjects at a sampling rate of 100\,Hz. We designate subjects 5 and 6 for inference, and use the rest for the database. Recordings are segmented into non-overlapping 2-second windows (200 time steps). For inference, we randomly sample 20 windows per activity from the inference users (10 from each) except for rope jumping, which has limited data. For this activity, we include 18 samples from subject 5 and 2 from subject 6, resulting in a total of 240 samples.

\paragraph{USC-HAD~\cite{10.1145/2370216.2370438}.} This dataset includes motion recordings from 14 subjects at a sampling rate of 100\,Hz. We designate subjects 13 and 14 for inference, using the remaining subjects to build the database. Data are segmented into non-overlapping 2-second windows (200 time steps). For inference, we randomly sample 20 windows per activity (10 from each subject), resulting in 240 total samples.

\paragraph{MHealth~\cite{Baos2014mHealthDroidAN}.} The MHealth dataset contains recordings from 10 subjects at a sampling rate of 50\,Hz. We use subjects 1 and 6 for inference and the remaining subjects to construct the database. Signals are segmented into non-overlapping 2-second windows (100 time steps). For inference, we randomly sample 20 windows per activity (10 from each subject), yielding a total of 240 evaluation samples.

\paragraph{WISDM~\cite{wisdm_smartphone_and_smartwatch_activity_and_biometrics_dataset__507}.} We use the smartwatch-on-hand subset of the WISDM dataset, recorded at 20\,Hz. Accelerometer and gyroscope signals are aligned by timestamp, and we select 47 users whose data show no alignment anomalies. Among them, 8 users are held out for inference and the rest are used to construct the database. Following the dataset’s recommendation, we segment the data into non-overlapping 10-second windows (200 time steps). For inference, we randomly sample 16 windows per activity (2 from each subject), resulting in a total of 288 inference samples.

\paragraph{DSADS~\cite{ALTUN20103605}.} The DSADS dataset contains recordings from 8 users at a sampling rate of 25\,Hz. We designate subjects 2 and 4 for inference and use the remaining users to build the database. We adopt the predefined 5-second windows (125 time steps) provided by the dataset. For inference, we randomly sample 10 windows per activity (5 from each subject), yielding a total of 190 inference samples.

\subsubsection{Cross-Dataset Generalization}
\label{apd_cross_dataset}
This section provides supplementary details for the cross-dataset generalization experiments in Section~\ref{exp:cross_dataset}, including dataset-specific preprocessing, activity-label mapping, and sensor-placement alignment used to establish comparable transfer scenarios. Following the common evaluation protocol defined in the main text, we restrict each transfer pair to the intersection of available sensor locations and activity classes, ensuring that all methods are evaluated under identical conditions.

Importantly, in the cross-dataset setting we transfer only the knowledge component, i.e., guidance on discriminative features, rather than retrieving evidence from the source domain. Because datasets differ in hardware, sampling rate, coordinate conventions, and wearing conditions, source-domain windows often carry dataset-specific signatures that do not reliably align with target-domain queries in the embedding space and may introduce spurious nearest-neighbor matches. We therefore construct the retrieval database exclusively from the target dataset to maintain evidence distribution alignment, while using cross-dataset knowledge to isolate and evaluate the transferability of motion priors.

\paragraph{UCI-HAR$\leftrightarrow$USC-HAD.}
This transfer pair evaluates cross-dataset generalization between UCI-HAR and USC-HAD over five shared activities: \emph{Walking, Walking upstairs, Walking downstairs, Sitting,} and \emph{Standing}. To satisfy the common evaluation protocol, we align sensor placement by mapping \emph{Waist} (UCI-HAR) to \emph{Front right hip} (USC-HAD). We additionally downsample USC-HAD to 50\,Hz to match the sampling rate of UCI-HAR. For cost-efficient inference with balanced coverage, we randomly sample 20 windows per activity from USC-HAD (10 per subject) and 40 windows per activity from UCI-HAR (4--5 per subject), ensuring non-overlapping instances.

\paragraph{PAMAP2$\leftrightarrow$WISDM.}
We evaluate transfer between PAMAP2 and WISDM on five overlapping activities: \emph{Walking, Standing, Jogging, Sitting,} and \emph{Stairs}. Sensor placements are aligned by mapping \emph{Wrist} (PAMAP2) to \emph{Hand} (WISDM). To harmonize temporal resolution, PAMAP2 is downsampled to 20\,Hz to match WISDM. We then construct balanced inference subsets by sampling 20 windows per activity from PAMAP2 (10 per subject) and 30 windows per activity from WISDM (3--4 per subject), with all samples non-overlapping.

\paragraph{MHealth$\leftrightarrow$WISDM.}
This pair studies transfer across MHealth and WISDM using the same five shared activities: \emph{Walking, Standing, Jogging, Sitting,} and \emph{Stairs}. We align sensor positions by mapping \emph{Right wrist} (MHealth) to \emph{Hand} (WISDM). Since WISDM is collected at 20\,Hz, MHealth is downsampled accordingly to ensure consistent input resolution. For inference, we randomly sample 20 windows per activity from MHealth (10 per subject) and 30 windows per activity from WISDM (3--4 per subject), maintaining balanced class and subject coverage with non-overlapping windows.

\paragraph{PAMAP2$\leftrightarrow$MHealth.}
For a multi-sensor transfer setting, we consider seven shared activities between PAMAP2 and MHealth: \emph{Lying, Sitting, Standing, Walking, Running, Cycling,} and \emph{Ascending stairs}. We align sensor placements across three locations using the following mappings: \emph{Wrist}$\leftrightarrow$\emph{Right wrist}, \emph{Chest}$\leftrightarrow$\emph{Chest}, and \emph{Ankle}$\leftrightarrow$\emph{Left ankle}. To match MHealth, PAMAP2 is downsampled to 50\,Hz. We then sample 20 non-overlapping windows per activity from each dataset (10 per subject), yielding balanced inference subsets for both domains.

\begin{table}[t]
\centering
\small
\resizebox{\linewidth}{!}{
\begin{tabular}{l
                cc cc
                cc cc
                cc cc
                cc}
\toprule
\multirow{2}{*}{\textbf{Dataset}} &
\multicolumn{2}{c}{\textbf{Agent 1}} &
\multicolumn{2}{c}{\textbf{Agent 2}} &
\multicolumn{2}{c}{\textbf{Agent 3}} &
\multicolumn{2}{c}{\textbf{Agent 4}} \\
\cmidrule(lr){2-3}\cmidrule(lr){4-5}\cmidrule(lr){6-7}\cmidrule(lr){8-9}
& \textbf{In} & \textbf{Out}
& \textbf{In} & \textbf{Out}
& \textbf{In} & \textbf{Out}
& \textbf{In} & \textbf{Out} \\
\midrule
Opportunity &1505  &343  &1784  &480  &941  &573  &1382  &781  \\
UCI-HAR     &2836  &354  &2651  &319  &604  &271  &802  &400  \\
Shoaib      &1815  &495  &3683  &304  &511  &220  & 711 &306  \\
PAMAP2      &13704  &1409  & 15664 &1033  &690  & 317 &913  & 454 \\
USC-HAD     & 8852 &1152  &14190  & 1547 &830 & 477 & 1296 &497  \\
MHealth     &2870  & 623 &7366  &537  & 451 & 206 &691  &271  \\
WISDM       & 8531 &1545  &15398  &1157  &610  &274  &809  &411  \\
DSADS       & 4834 & 1036 & 9695 &847  &606  & 314 & 870 &389  \\
\bottomrule
\end{tabular}
}
\caption{Per-dataset token usage (input/output) for each agent stage. All token statistics are collected from experiments using Gemini-2.0-Flash for all agent stages.}
\label{tab:token_usage}
\end{table}

\begin{table}[t]
\centering
\small
\resizebox{\linewidth}{!}{
\begin{tabular}{lcccc}
\toprule
\textbf{Dataset} &
\multicolumn{2}{c}{\textbf{With Retrieval}} &
\multicolumn{2}{c}{\textbf{No Retrieval}} \\  \cmidrule(lr){2-3} \cmidrule(lr){4-5}
& \textbf{Acc} & \textbf{UB} & \textbf{Acc} & \textbf{UB} \\
\midrule
Opportunity & 92.5 & 96.0 & 84.0 & 92.0 \\
UCI-HAR     & 90.0 & 99.6 & 86.3 & 99.2 \\
Shoaib      & 97.1 & 99.5 & 91.4 & 99.5 \\
PAMAP2      & 76.7 & 84.2 & 57.9 & 72.1 \\
USC-HAD     & 60.0 & 80.8 & 47.9 & 71.3 \\
MHealth     & 86.3 & 99.6 & 76.3 & 98.3 \\
WISDM       & 65.6 & 78.8 & 54.9 & 75.0 \\
DSADS       & 84.2 & 92.6 & 75.3 & 85.8 \\
\midrule
\textbf{Average} & \textbf{81.6} & \textbf{91.4} & \textbf{71.8} & \textbf{86.7} \\
\bottomrule
\end{tabular}
}
\caption{Impact of Retrieval on \textit{Evidence Pruning} and \textit{Decision and Insight} Agents. We report training-free accuracy (Acc) and upper-bound accuracy (UB) with and without retrieval across all datasets.}
\label{tab:retrieval_ablation}
\vspace{-1em}
\end{table}

\begin{table}[t]
\centering
\small
\resizebox{\linewidth}{!}{
\begin{tabular}{lcccc}
\toprule
\textbf{Dataset} & 
\makecell[c]{\textbf{Pruning}\\\textbf{}} & 
\makecell[c]{\textbf{Upper}\\\textbf{Bound}} & 
\makecell[c]{\textbf{No}\\\textbf{Pruning}} & 
\makecell[c]{\textbf{Avg.}\\\textbf{Length}} \\
\midrule
Opportunity & 92.5 & 96.0 & 73.0 & 2.04 \\
UCI-HAR     & 90.0 & 99.6 & 78.3 & 2.20 \\
Shoaib      & 97.1 & 99.5 & 93.3 & 2.42 \\
PAMAP2      & 76.7 & 84.2 & 55.8 & 2.58 \\
USC-HAD     & 60.0 & 80.8 & 50.4 & 2.86 \\
MHealth     & 86.3 & 99.6 & 76.7 & 2.59 \\
WISDM       & 65.6 & 78.8 & 54.2 & 2.56 \\
DSADS       & 84.2 & 92.6 & 64.2 & 2.65 \\
\midrule
\textbf{Average} & \textbf{81.6} & \textbf{91.4} & \textbf{68.2} & \textbf{2.49} \\
\bottomrule
\end{tabular}
}
\caption{Impact of the Evidence Pruning Agent: Accuracy (\%) and Upper Bound with and without pruning, along with the average pruned shortlist length per dataset.}
\label{tab:pruning-ablation}
\vspace{-0.5em}
\end{table}

\subsection{Token Usage Breakdown}
\label{apd_token_usage}
To quantify the inference cost of ZARA, we report the token usage of each agent stage across all datasets. For every dataset, we measure the input and output tokens consumed by each agent and report the averages in Table~\ref{tab:token_usage}. This analysis complements the main results by making the cost--accuracy trade-offs explicit and enabling transparent comparison of compute overhead across evaluation settings.

As shown in Table~\ref{tab:token_usage}, token consumption is dominated by the early agent stages responsible for candidate construction and for assembling knowledge and retrieval evidence, whereas the later reasoning stages are comparatively lightweight. Across datasets, larger input-token counts generally correlate with richer sensor knowledge and a larger number of activity classes, both of which increase the amount of numerical evidence and class-conditioned comparisons included in the prompt. In all cases, input tokens consistently exceed output tokens, indicating that inference cost is driven more by contextual grounding than by generation length. Importantly, our prompts remain substantially shorter than approaches that directly linearize raw sensor signals into text, whose token cost scales with sequence length and channel count and thus quickly becomes impractical in API-based inference. Future work could further reduce inference cost by caching static knowledge, compressing numerical evidence, or using smaller candidate sets.

\subsection{Ablation}
\label{apd_ablation}
This section provides detailed results for the ablation studies presented in Section~\ref{ablation}. All results are obtained using Gemini-2.0-Flash.

\paragraph{Removing Retrieval Reduces Performance.}
Table~\ref{tab:retrieval_ablation} reports the exact values underlying Figure~\ref{fig:retrieval_ablation}, comparing ZARA with and without the Evidence Retrieval module. We report both the final accuracy in the training-free setting and the pruning-stage upper-bound accuracy for each dataset.

\paragraph{Skipping Evidence Pruning Hurts.}
Table~\ref{tab:pruning-ablation} reports the dataset-level breakdown of ZARA's accuracy in the training-free setting, with and without the Evidence Pruning Agent, together with the corresponding upper-bound accuracy and the average length of the pruned candidate shortlist. These results complement Figure~\ref{fig:prune} in the main paper and confirm that pruning substantially improves performance while preserving high-quality candidates.

\begin{table}[t]
\centering
\small
\resizebox{\linewidth}{!}{
\begin{tabular}{lcccc}
\toprule
\textbf{Dataset} &
\multicolumn{2}{c}{\textbf{With knowledge}} &
\multicolumn{2}{c}{\textbf{No knowledge}} \\  \cmidrule(lr){2-3} \cmidrule(lr){4-5}
& \textbf{Acc} & \textbf{UB} & \textbf{Acc} & \textbf{UB} \\
\midrule
Opportunity  & 92.5  & 96.0 & 82.0 & 99.0 \\
UCI-HAR      & 90.0  & 99.6 & 68.8 & 98.9 \\
Shoaib       & 97.1  & 99.5 & 77.6 & 97.1 \\
PAMAP2       & 76.7  & 84.2 & 56.7 & 71.3 \\
USC-HAD      & 60.0  & 80.8 & 35.4 & 81.7 \\
Mhealth      & 86.3  & 99.6 & 80.8 & 94.6 \\
WISDM        & 65.6  & 78.8 & 53.8 & 73.6 \\
DSADS        & 84.2  & 92.6 & 52.1 & 79.5 \\
\midrule
\textbf{Average} & \textbf{81.6} & \textbf{91.4} & \textbf{63.4} & \textbf{87.0} \\
\bottomrule
\end{tabular}
}
\caption{Impact of Prior Knowledge Injection on Feature Selector Accuracy and Upper Bound.}
\label{tab:knowledge-table}
\end{table}

\begin{table}[t]
\centering
\small
\resizebox{\linewidth}{!}{
\begin{tabular}{lccccccc}
\toprule
\textbf{Dataset} & \makecell{\textbf{\# of} \\ \textbf{Channels}} & \makecell{\textbf{Window} \\ \textbf{Size}} & \makecell{\textbf{Database} \\ \textbf{Size}} & \textbf{DTW} & \textbf{Moment-s} & \textbf{Moment-l} & \textbf{Mantis} \\
\midrule
Opportunity  & 30 & 60  & 6968  & 0.5814 & 0.0683 & 0.1508 & 0.3072 \\
UCI-HAR      & 6  & 128 & 7352  & 0.1389 & 0.0169 & 0.0342 & 0.0623 \\
Shoaib       & 30 & 100 & 5040  & 0.4935 & 0.0686 & 0.1700 & 0.3122 \\
PAMAP2       & 18 & 200 & 7138  & 0.5104 & 0.0446 & 0.1053 & 0.1777 \\
USC-HAD      & 6  & 200 & 11889 & 0.2584 & 0.0180 & 0.0389 & 0.0699 \\
Mhealth      & 15 & 100 & 2799  & 0.1504 & 0.0405 & 0.0900 & 0.1528 \\
WISDM        & 6  & 200 & 14287 & 0.3152 & 0.0190 & 0.0369 & 0.0679 \\
DSADS        & 30 & 125 & 6840  & 0.6127 & 0.0741 & 0.1762 & 0.3105 \\
\bottomrule
\end{tabular}
}
\caption{Per-query retrieval latency (in seconds) and dataset characteristics.}
\label{tab:retrieval-latency}
\vspace{-0.5em}
\end{table}

\paragraph{No Prior Knowledge Fails.}
Table~\ref{tab:knowledge-table} reports the exact values plotted in Figure~\ref{fig:knowledge_ablation}, comparing ZARA with and without the prior knowledge base across all eight datasets. These results confirm that prior knowledge plays a critical role in both narrowing the candidate set and distinguishing among activity classes.

\paragraph{Retrieval Latency by Dataset.}
Table~\ref{tab:retrieval-latency} reports the average per-query latency (in seconds) of each retrieval method across all eight datasets. Measurements were taken on an Apple M2 Max CPU with 64GB memory. Although latency varies with window length, number of channels, and database size, the relative ranking remains consistent: DTW is the slowest, Moment-Small is the fastest, and Mantis offers a balanced trade-off between speed and retrieval quality.

\begin{figure*}[t]
\centering
\includegraphics[width=1\linewidth]{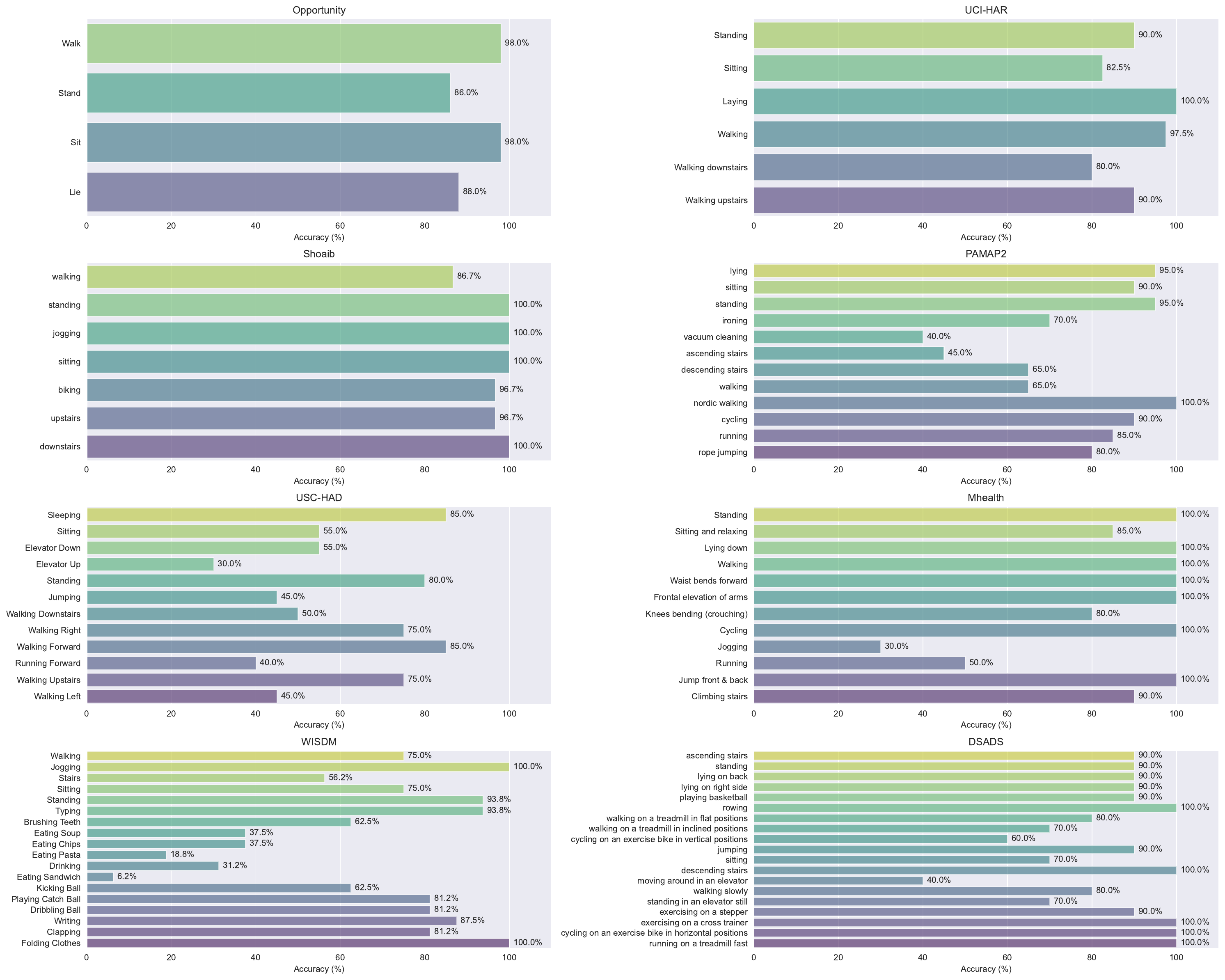} 
\caption{Per-Class Evaluation.}
\label{fig:per-class-results}
\end{figure*}

\paragraph{Per-Class Evaluation.}
Table~\ref{tab:zeroshot} in the main paper reports only the overall accuracy and macro F1 for each dataset. To further assess performance consistency across activity types, we provide detailed per-class accuracy in Figure~\ref{fig:per-class-results}. ZARA consistently achieves strong per-class performance, reflecting its ability to distinguish a wide range of behaviors through structured knowledge and retrieved evidence. By contrast, several baselines exhibit substantially lower F1 than accuracy, indicating a bias toward dominant classes. This further highlights ZARA's stronger and more balanced generalization across HAR activities.

\subsection{Predefined Features}
\label{apd_feats}
To support feature-based reasoning and retrieval, we extract a comprehensive set of handcrafted features from each sensor channel (6 axes per sensor plus 2 magnitude channels). These features span both the time and frequency domains, and are designed to capture fine-grained temporal, statistical, and spectral characteristics of motion signals. The full feature set used for each channel is summarized in Table~\ref{tab:feature_list}.

\begin{table*}[t]
\centering
\small
\resizebox{\textwidth}{!}{
\begin{tabular}{p{0.22\textwidth} p{0.72\textwidth}}
\toprule
\textbf{Category} & \textbf{Feature Description} \\
\midrule
Time-domain Features &
Mean, Standard Deviation (STD), Variance, Maximum, Minimum, Median, Root Mean Square (RMS), 
Peak Amplitude, Zero-Crossing Rate, Slope, 
Mean/RMS/STD of First-Order Differences, Range, Sum, 
Signal Absolute Value, Mean Absolute Value, 
Interquartile Range, Skewness, Kurtosis, 
Signal Magnitude Area \\
\midrule
Frequency-domain (FFT) &
Band Power (Low, Mid, High), Band Power Ratio (Low, Mid, High), 
Dominant Frequency, Power of Dominant Frequency, 
Second Peak Frequency and Power, 
Spectral Centroid, Spectral Entropy, 
Spectral Skewness, Spectral Kurtosis, 
Weighted Average Frequency, Spectral Energy, Max Power Index \\
\midrule
Frequency-domain (STFT) &
STFT Max / Mean / STD in Low, Mid, High bands, 
STFT Entropy (Mean, Max, STD), 
STFT Centroid (Mean, Max, STD) \\
\midrule
Autocorrelation & 
First Peak Lag, First Minimum Lag, First Zero-Crossing Lag \\
\midrule
Jerk-based Features & 
Jerk RMS, Peak, Zero-Crossing Rate \\
\midrule
Cross-Channel Features & 
Pearson Correlation Between Channels \\
\bottomrule
\end{tabular}
}
\caption{Predefined features extracted per sensor channel (6 axes + 2 magnitudes per sensor).}
\label{tab:feature_list}
\end{table*}

\subsection{Prompt Template}
\label{apd_prompt}
Below we provide the full prompts used by ZARA during inference. These prompts cover all stages of the reasoning pipeline: the First Feature Selector (Figure~\ref{fig:choose_features_prompt}), Evidence Pruning (Figure~\ref{fig:evidence_prompt}), the Second Feature Selector (Figure~\ref{fig:feature_prompt2}), and Decision Insight (Figure~\ref{fig:decision_prompt}), together with their corresponding inputs and output formats. Although the exact wording may vary slightly across datasets or instances, the overall structure is consistent throughout all experiments.

\begin{figure*}[t]
\centering
\scalebox{0.92}{ 
\begin{minipage}{\textwidth} 
    \begin{tcolorbox}[width=1\textwidth,colback=gray!5!white,colframe=black!75!white,title=System Instruction]
    \textbf{Task} \\
    You're an expert in Human Activity Recognition, with a focus on identifying the most effective features for distinguishing between human activities. \\
    
    \textbf{Glossary of Abbreviations} \\
    \texttt{{GLOSS\_TEXT}} \\
    
    \textbf{Instructions}
    \begin{itemize}
      \item Based on the user-provided “Top Features per Activity Pair“, select up to \texttt{{TOP\_N}} unique features that best distinguish the specified Target Activities.
      \item When selecting features, prioritize those that:
      \begin{itemize}
        \item Appear consistently across multiple activity pairs, or
        \item Have relatively high importance scores within specific pairs.
      \end{itemize}
    \end{itemize}

    \textbf{Output Format} \\
    Return \textbf{only} the following, with no extra text or line breaks:
    
    \begin{center}
    \begin{tabular}{|c|c|}
    \hline
    \textbf{Index} & \textbf{Feature Name} \\
    \hline
    \end{tabular}
    \end{center}
    \end{tcolorbox}
    
    \begin{tcolorbox}[width=1\textwidth,colback=gray!5!white,colframe=black!75!white,title=User Prompt]
    \textbf{Target Activities} \\
    \texttt{{ACTIVITY\_LIST}}   \\
    
    \textbf{Top Features per Activity Pair (Ranked by Importance Score)} \\
    (\texttt{{PAIR\_NUM}} activity pairs in total) 
    
    \texttt{{PAIR\_WISE\_KNOWLEDGE}}
    \end{tcolorbox}
\end{minipage}
}
\caption{Prompt template for the first Feature Selector agent.}
\label{fig:choose_features_prompt}
\end{figure*}

\begin{figure*}[t]
\centering
\scalebox{0.92}{ 
\begin{minipage}{\textwidth} 
    \begin{tcolorbox}[width=1\textwidth,colback=gray!5!white,colframe=black!75!white,title=System Instruction]
    \textbf{Task} \\
    You're an expert in Human Activity Recognition. Your task is to narrow down the set of plausible activities for the \textbf{QUERY} segment, based on the given features and their statistical distributions in the user-supplied activities table. \\
    
    \textbf{Instructions}
    \begin{itemize}
      \item Each row in the activities table represents an activity class, with cells showing the mean ± std of each feature.
      \item The QUERY row presents the feature values of the segment to classify.
      \item You must select at least two activity classes, and ideally all activity classes that are reasonably or even marginally plausible for the QUERY.
        \item For each selected activity class, provide a brief explanation of why it is a plausible match.
    \end{itemize}
    
    \textbf{Output Format} \\
    Return \textbf{only} the following, in this exact order, with no additional text or line breaks:
    
    \begin{center}
    \begin{tabular}{|c|c|c|}
    \hline
    \textbf{Index} & \textbf{Activity} & \textbf{Reason} \\
    \hline
    \end{tabular}
    \end{center}
    \end{tcolorbox}
    
    \begin{tcolorbox}[width=1\textwidth,colback=gray!5!white,colframe=black!75!white,title=User Prompt]
    \textbf{Activities Table} \\
    \texttt{{ACTIVITIES\_FEATURES\_TABLE}}   
    \end{tcolorbox}
\end{minipage}
}
\caption{Prompt template for Evidence Pruning agent.}
\label{fig:evidence_prompt}
\end{figure*}

\begin{figure*}[t]
\centering
    \begin{tcolorbox}[width=1\textwidth,colback=gray!5!white,colframe=black!75!white,title=System Instruction]
    \textbf{Task} \\
    You're an expert in Human Activity Recognition, with a focus on identifying the most effective features for distinguishing between human activities. \\
    
    \textbf{Glossary of Abbreviations} \\
    \texttt{{GLOSS\_TEXT}} \\
    
    \textbf{Instructions}
    \begin{itemize}
      \item Based on the user-provided “Top Features per Activity Pair“, select up to \texttt{{TOP\_N}} unique features that best distinguish the specified Target Activities.
      \item When selecting features, prioritize those that:
      \begin{itemize}
        \item Appear consistently across multiple activity pairs, or
        \item Have relatively high importance scores within specific pairs.
      \end{itemize}
      \item For each selected feature, give:
      \begin{itemize}
        \item Definition – A concise, clear explanation of the feature.
        \item Discriminative Power – Summarize the following:
            \begin{itemize}
            \item Which activity pairs this feature helps to distinguish.
            \item The relative importance rate of this feature within each activity pair, indicating how effectively it differentiates between the two activities in that pair.
          \end{itemize}
      \end{itemize}
    \end{itemize}
    
    \textbf{Output Format} \\
    Return \textbf{only} the following, with no extra text or line breaks:
    
    \begin{center}
    \begin{tabular}{|c|c|c|c|}
    \hline
    \textbf{Index} & \textbf{Feature Name} & \textbf{Definition} & \textbf{Discriminative Power} \\
    \hline
    \end{tabular}
    \end{center}
    \end{tcolorbox}
    
    \begin{tcolorbox}[width=1\textwidth,colback=gray!5!white,colframe=black!75!white,title=User Prompt]
    \textbf{Target Activities} \\
    \texttt{{ACTIVITY\_LIST}}   \\
    
    \textbf{Top Features per Activity Pair (Ranked by Importance Score)} \\
    (\texttt{{PAIR\_NUM}} activity pairs in total) 
    
    \texttt{{PAIR\_WISE\_KNOWLEDGE}}
    \end{tcolorbox}

\caption{Prompt template for second Feature Selector agent.}
\label{fig:feature_prompt2}
\end{figure*}

\begin{figure*}[t]
\centering
    \begin{tcolorbox}[width=1\textwidth,colback=gray!5!white,colframe=black!75!white,title=System Instruction]
    \textbf{Task} \\
    You are an expert in Human Activity Recognition. Your goal is to determine the \textbf{most probable activity class} for the \textbf{QUERY} segment by comparing its feature values against the statistical distributions in the user-provided activities table. \\
    
    \textbf{Sensor Feature Explanation Guide Table}\\
    This table describes each feature and indicates which activity classes it helps to distinguish between.\\
    \texttt{{FEATURES\_REFERENCE\_TABLE}} \\
    
    \textbf{Instructions}
    \begin{itemize}
      \item Each row in the activities table corresponds to an activity class, with each cell showing the mean ± standard deviation for a feature.
      \item The QUERY row presents the feature values of the segment to classify.
      \item Select the single most likely activity class, and base your decision on specific feature(s) in the QUERY row.
        \item In your explanation:
            \begin{itemize}
                \item Explicitly compare the Query’s feature values to each class’s distribution, explaining why the predicted class is a better match than each alternative.
                \item When unsure, refer to the Discriminative Power in the guide table to justify how strongly each feature helps distinguish the specific activities.
            \end{itemize}
    \end{itemize}
    
    \textbf{Output Format} \\
    Respond with \textbf{exactly one} line in this JSON format (no extra text or line breaks):
    \begin{center}
    \begin{verbatim}
    ```json
    {
      "reason": "<your detailed explanation>",
      "predicted_class": "<ClassName>"
    }
    \end{verbatim}
    \end{center}
    \end{tcolorbox}
    
    \begin{tcolorbox}[width=1\textwidth,colback=gray!5!white,colframe=black!75!white,title=User Prompt]
    \textbf{Activities Table} \\
    \texttt{{ACTIVITIES\_FEATURES\_TABLE}}   
    \end{tcolorbox}
\caption{Prompt template for Decision Insight agent.}
\label{fig:decision_prompt}
\end{figure*}

\end{document}